\documentclass[10pt,twocolumn,letterpaper]{article}

\usepackage{iccv}
\usepackage{times}
\usepackage{epsfig}
\usepackage{graphicx}
\usepackage{amsmath}
\usepackage{amssymb}
\usepackage{booktabs}
\usepackage{gensymb}

\usepackage[dvipsnames,table]{xcolor}

\newcommand{\plus}[1]{\textcolor{blue}{\textbf{\scriptsize{(+#1)}}}}
\newcommand{\minus}[1]{\textcolor{red}{\textbf{\scriptsize{(-#1)}}}}

\usepackage[dvipsnames]{xcolor}

\usepackage[pagebackref=true,breaklinks=true,letterpaper=true,colorlinks,bookmarks=false,citecolor=ForestGreen]{hyperref}

\iccvfinalcopy 

\def\iccvPaperID{9639} 

\ificcvfinal\fi

\begin{document}

\title{Multi-Task Self-Training for Learning General Representations}

\newcommand{\aspace}{\enspace}
\author{Golnaz Ghiasi\thanks{Authors contributed equally.}, Barret Zoph\footnotemark[1],\aspace Ekin D. Cubuk\footnotemark[1],\aspace Quoc V. Le,\aspace Tsung-Yi Lin \\
Google Research, Brain Team \\
{\tt\small \{golnazg,barretzoph,cubuk,qvl,tsungyi\}@google.com}}

\maketitle
\ificcvfinal\thispagestyle{empty}\fi

\begin{abstract}
Despite the fast progress in training specialized models for various tasks, learning a single general model that works well for many tasks is still challenging for computer vision. 
Here we introduce multi-task self-training (MuST), which harnesses the knowledge in  independent specialized teacher models (\eg, ImageNet model on classification) to train a single general student model. Our approach has three steps. First, we train  specialized teachers independently on labeled datasets. We then use the specialized teachers to label an unlabeled dataset to create a multi-task pseudo labeled dataset. Finally, the dataset, which now contains pseudo labels from teacher models trained on different datasets/tasks, is then used to train a student model with multi-task learning. We evaluate the feature representations of the student model on 6 vision tasks including image recognition (classification, detection, segmentation) and 3D geometry estimation (depth and surface normal estimation). MuST is scalable with unlabeled or partially labeled datasets and outperforms both specialized supervised models and self-supervised models when training on large scale datasets. Lastly, we show MuST can improve upon already strong checkpoints~\cite{chao2021align} trained with billions of examples. The results suggest self-training is a promising direction to aggregate labeled and unlabeled training data for learning general feature representations.

\end{abstract}

\section{Introduction}

Visual representation learning is a core problem in computer vision. Supervised and self-supervised pre-training have shown promising results in transferring the learned feature representations to downstream tasks. Typically, a model is pre-trained with a supervised~\cite{kolesnikov2020big,dosovitskiy2020image} or a self-supervised objective~\cite{chen2020simple,grill2020bootstrap,he2019momentum}. Despite the wide adoption of transfer learning from supervised training, the features may not necessarily be useful for downstream tasks. 
For example, He \etal found that ImageNet pre-training fails to improve COCO instance segmentation~\cite{rethinking}. In contrast, Shao \etal showed features learned from Objects365 detection dataset improve COCO instance segmentation by a large margin~\cite{objects365}. Pre-training with a specialized task that aligns with the downstream target task still yields the best performance in object detection~\cite{li2019detection_pretraining,objects365} and semantic segmentation~\cite{chen2017deeplab}.

\begin{figure}[t!]
    \centering
    \includegraphics[width=0.45\textwidth]{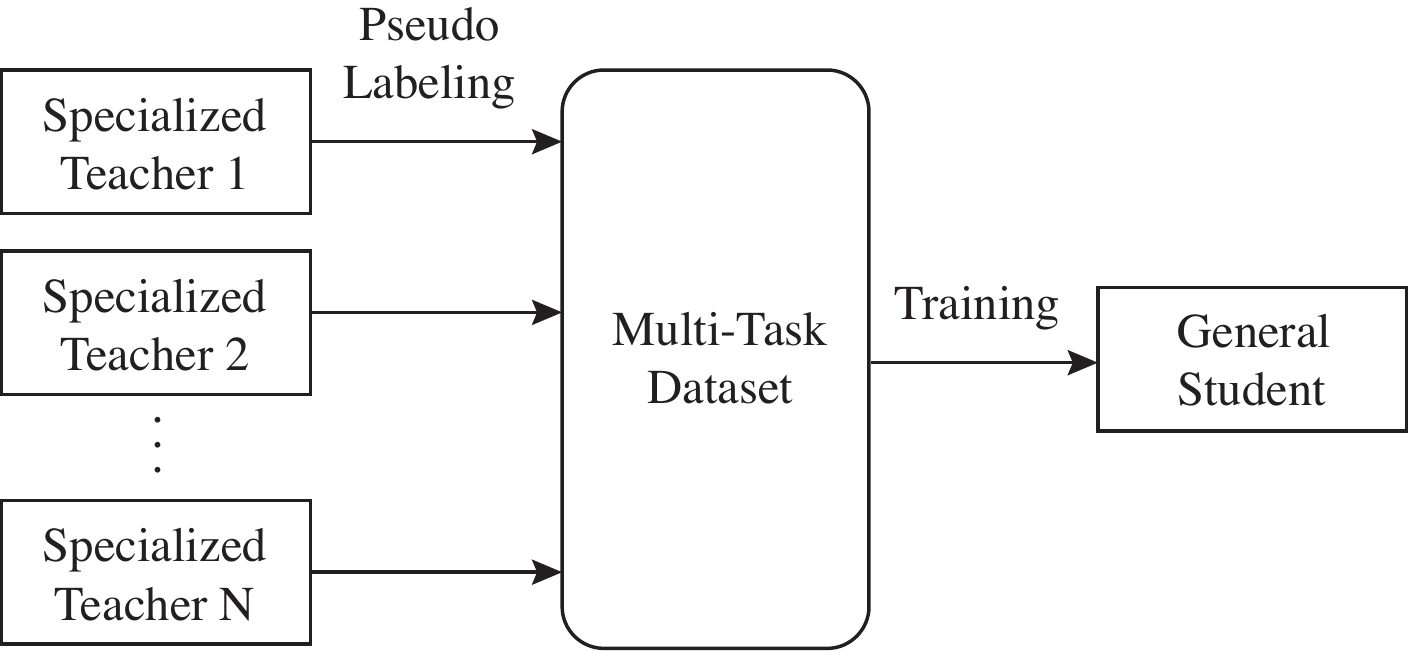}
    \caption{\textbf{An overview of Multi-Task Self-Training (MuST).} Specialized Teacher represents a supervised model trained on a single task and dataset (\eg, classification model trained on ImageNet). Specialized Teacher models are trained independently on their own tasks and datasets. They then generate pseudo labels on a shared dataset. Finally, a single General Student model is trained jointly using the pseudo (and supervised) labels on the shared dataset.}
    \label{fig:concept}
\end{figure}

Intuitively, it is possible to learn general features by training a model to simultaneously do well on multiple tasks. Recent work in NLP started to show promising results on learning a generalist model with multi-task learning~\cite{xue2020mt5,clark2019bam}. In computer vision, the biggest challenge of training a multi-task model is in the data collection and annotation. Despite datasets like COCO~\cite{coco}, collecting a wide variety of annotations (\eg, instance segmentation, person keypoints, image caption) for the same image dataset is quite challenging. Due to the time consuming nature of annotating images with labels, it is hard to scale such efforts with the number of images and the number of tasks. The lack of large scale multi-task datasets impedes the progress in multi-task learning for computer vision.

\begin{figure}[t!]
    \centering
    \includegraphics[width=0.325\linewidth]{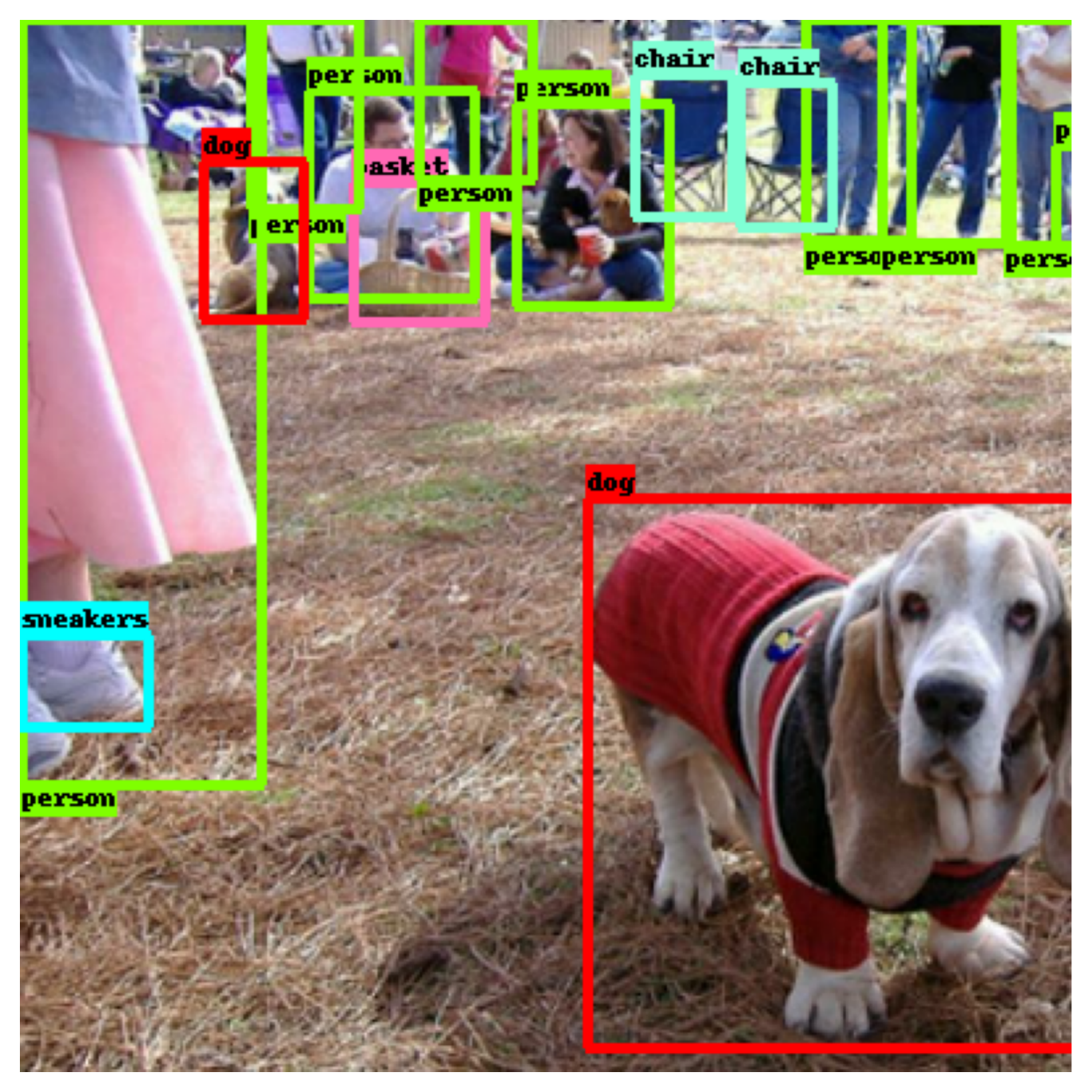}
    \includegraphics[width=0.325\linewidth]{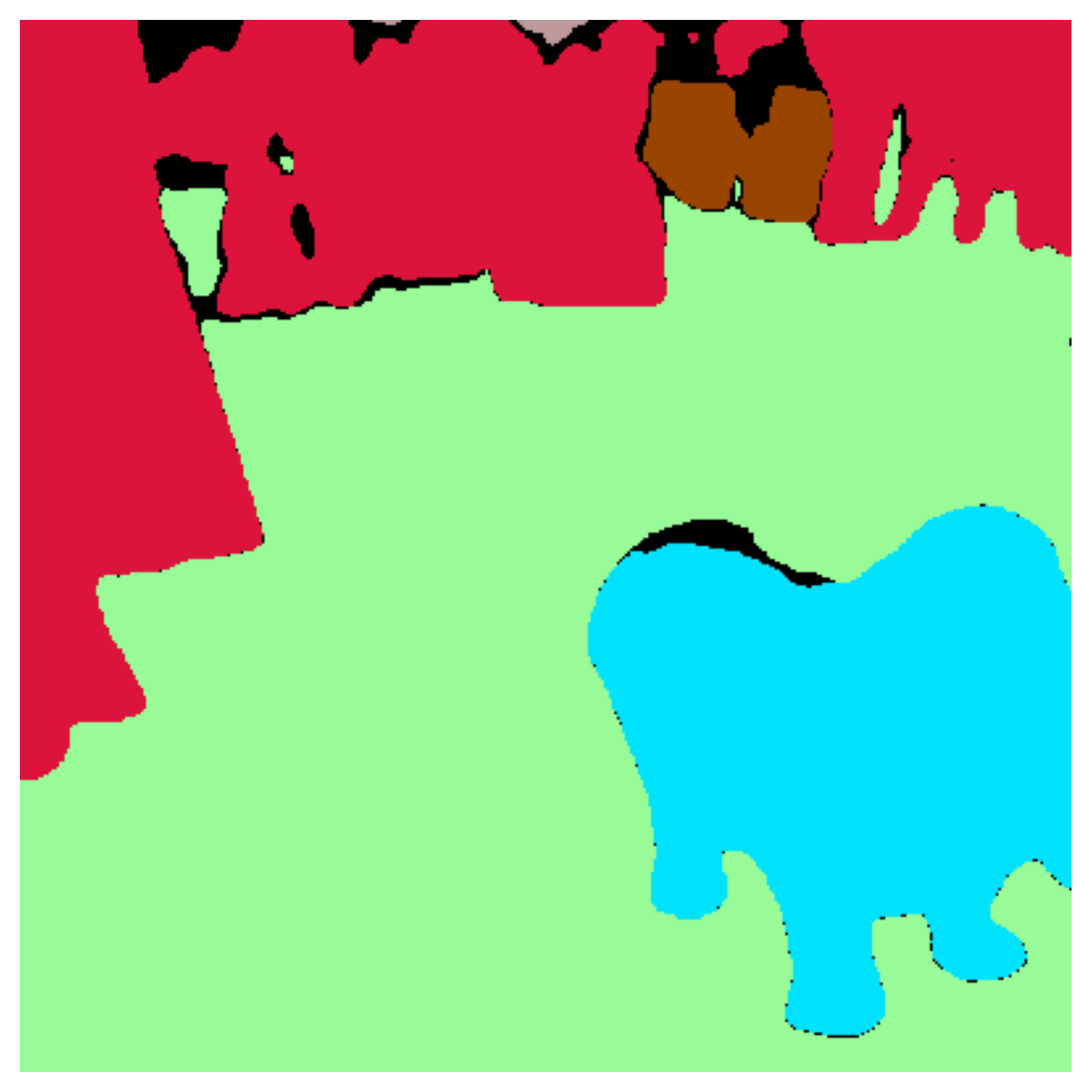}
    \includegraphics[width=0.325\linewidth]{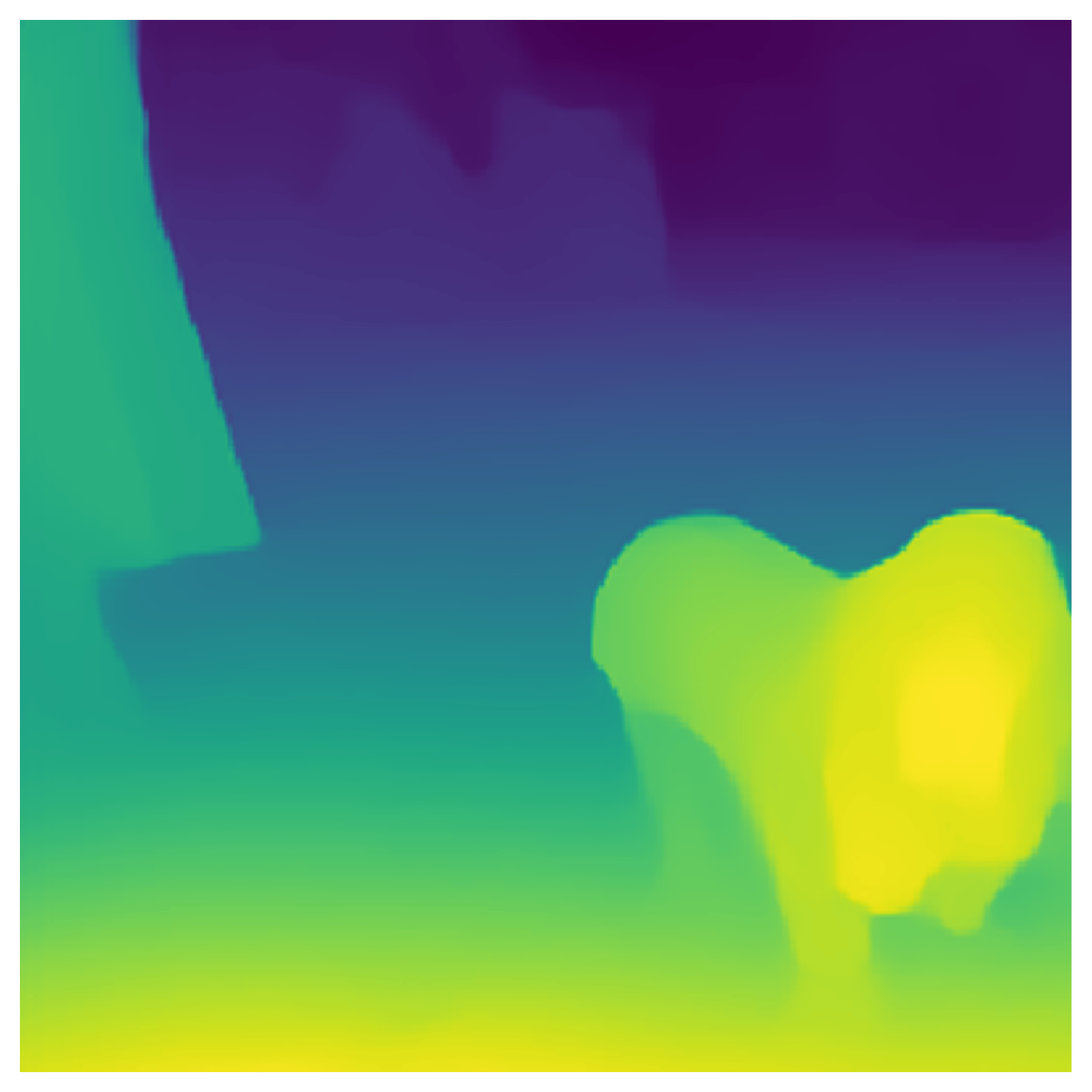}
    \caption{\textbf{Examples of pseudo labels on ImageNet.} \textbf{Left:} bounding boxes labeled with an Objects365 teacher model. \textbf{Middle:} semantic segmentation labeled with a COCO teacher model. \textbf{Right:} depth labeled with a MiDaS teacher model.}
    \label{fig:pseudo_labels}
\end{figure}

In this work, we study using self-training to remedy the issue. We propose to use pseudo labeling to enable large scale multi-task feature learning for computer vision. Zoph \etal~\cite{zoph2020rethink} observed that self-training further improves pre-training for transfer learning, and that self-training works even when pre-training fails to outperform a randomly initialized model. The gap between pre-training and self-training suggests that self-training can learn better features from pseudo labels. Inspired by this observation, we first investigate whether good features can be learned by only using pseudo labels. We train teacher models using datasets such as COCO or Objects365 to generate pseudo labels on unlabeled images. Figure~\ref{fig:pseudo_labels} shows example pseudo labels on ImageNet. Surprisingly, we find a student model trained with only these pseudo labels preserves most of the transfer learning performance of its specialized teacher model.
This finding suggests pseudo labels are effective at distilling the knowledge in a supervised dataset. Therefore, we can use pseudo labels to transfer knowledge from multiple teacher models to a single student model for representation learning.

We propose Multi-Task Self-Training (MuST) to train a generalist student model on the information distilled from teacher models trained on different tasks and datasets. Figure~\ref{fig:concept} shows the overview of the algorithm. MuST has three steps. First, it trains specialized teachers independently on labeled datasets. For example, one teacher can be trained with depth prediction and another teacher can be trained with object detection. The specialized teachers are then used to label a larger unlabeled dataset to create a multi-task pseudo labeled dataset. For example, these teachers can generate depth estimations and object detections on the ImageNet dataset. Finally, the dataset, which now contains pseudo labels from teacher models trained on different datasets/tasks, is used to train a student model with multi-task learning. Hence the student, for example, can do depth prediction and object detection at the same time.

\begin{figure}[t!]
    \centering
    \includegraphics[width=0.45\textwidth]{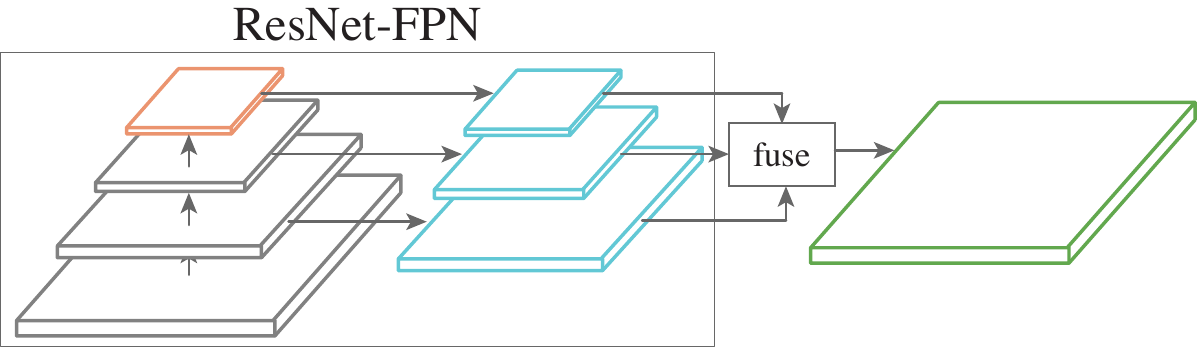}
    \caption{\textbf{The ResNet-FPN backbone architecture for multi-task learning.} \textcolor{orange}{\textbf{Orange:}} the top-level features for classification. \textcolor{cyan}{\textbf{Cyan:}} multi-scale features for box detection and instance segmentation. \textcolor{LimeGreen}{\textbf{Green:}} the high resolution features for pixel-wise tasks (\eg, segmentation, depth, and surface normal estimation.)}
    \label{fig:model}
\end{figure}

In our experiments, we have four teacher models: classification, semantic segmentation, object box detection, and depth estimation. We design a simple model architecture (Figure~\ref{fig:model}) based on ResNet~\cite{resnet} and feature pyramid networks (FPN)~\cite{fpn}. The parameters in the ResNet-FPN backbone are shared across different tasks. For each individual task, it has a small task-specific head consisting of a few convolution layers followed by a linear prediction layer. Our experiments show that this simple model architecture is able to absorb the knowledge of different tasks in the shared backbone. The generalist student model is on par with/outperforms its specialist teacher models for all transfer learning tasks.

The recent self-supervised algorithms like SimCLR~\cite{chen2020simple}, MoCo~\cite{he2019momentum} are shown to create representations that are on par or better than its supervised counterpart.
In our experiments, MuST also outperforms SimCLR~\cite{chen2020simple} by a large margin on segmentation and depth estimation tasks.
We also observe that the representations learned by SimCLR is on par with those of supervised learning on ImageNet (1.3M images) but does not scale as well on JFT (300M images). On the contrary, MuST outperforms SimCLR~\cite{chen2020simple} on both ImageNet and JFT. 
Moreover, MuST also outperforms supervised JFT pre-training for 5 out of 6 tasks except the image classification task. The results indicate the potential of MuST in learning general feature representations that improve with more unlabeled data.

Lastly, we show MuST can improve upon already strong checkpoints such as ALIGN~\cite{chao2021align}. We fine-tune ALIGN checkpoints, previously trained with billions of supervised examples, with MuST pseudo labels and  find improvements on a suite of downstream tasks: detection, segmentation, and depth estimation tasks.

We summarize our contributions below:
\begin{itemize}
    \item We propose Multi-Task Self-Training (MuST), a simple algorithm for creating general visual representations by multi-task learning with pseudo labels.
    \item  We conduct experiments by jointly training across several datasets (\eg, ImageNet, Objects365, COCO, JFT) to learn general feature representations that outperforms representations learned by supervised and self-supervised methods. 
    \item We perform experiments to compare supervised, self-supervised, and MuST on 6 computer vision tasks including tasks in image recognition (classification, detection, segmentation) and 3D geometry estimation (depth and surface normal estimation).
    \item MuST can be used to improve upon already strong checkpoints and achieve competitive results on a variety of tasks compared to task-specific state-of-the-art models.
\end{itemize}

\section{Related Work}
\paragraph{Multi-Task learning: } Multi-task learning has a rich history in deep learning~\cite{ruder2017overview}. A common strategy for multi-task learning is to share the hidden layers of a ``backbone'' model for different tasks~\cite{caruana1997multitask}. More recently, multi-task learning has led to improved accuracy in NLP~\cite{clark2019bam,gao2019multitasknlp}. Although, Raffel~\etal found that multi-task learning generally underperformed compared to pre-training followed by fine-tuning~\cite{raffel2019exploring}.

In the vision domain, Zamir~\etal studied the transfer learning dependencies across 26 tasks with an indoor dataset~\cite{zamir2018taskonomy}. Instead of exploring the task dependencies, we are interested in pushing a single model that can absorb knowledge of all tasks for learning general representations. Kokkinos~\etal~\cite{Kokkinos_2017_CVPR} and Xiao\etal~\cite{xiao2018unified} trained models across multiple datasets by simply zeroing losses for examples that don't have labels for a particular task. We propose to apply pseudo labels so every image is annotated with all tasks. Girshick~\etal used a multi-task loss for classification and bounding-box regression to improve the training of object detectors~\cite{girshick2015fast}. We follow the similar approach of using one large backbone model and smaller heads for multiple tasks.

\paragraph{Self-training:} Self-training is a popular technique to incorporate unlabeled data into supervised learning~\cite{yarowsky1995unsupervised,scudder1965probability,rosenberg2005semi,lee2013pseudo}. The method works by using a supervised model to generate pseudo labels on unlabeled data. Then a student model is  trained on the pseudo labeled data.
Yalniz \etal~\cite{zeki2019billion} showed a model ``pre-trained'' with pseudo labels on a large unlabeled dataset (at hundreds millions scale) can improve classification accuracy. 
Noisy Student~\cite{xie2019self} used self-training to push state-of-the-art performance on ImageNet by training jointly with 130M pseudo labeled images.
Chen \etal~\cite{chen2020naive} obtained state-of-the-art panoptic segmentation results on Cityscapes with self-training.
Zoph \etal~\cite{zoph2020rethink} improved the state-of-the-art on object detection and semantic segmentation with self-training.
All the above works focused on a single task. On the contrary, our work focuses on using self-training for multi-task learning to learn general representations.

\paragraph{Representation learning:} 
Transfer learning from ImageNet pre-training has been the most widely used method in computer vision.
BiT~\cite{kolesnikov2020big} and ViT~\cite{dosovitskiy2020image} pre-trained the model on JFT-300M dataset~\cite{sun2017revisiting} and obtained strong performance when fine-tuned on downstream vision tasks. In particular, Mahajan \etal showed model pre-trained with Instagram benefits other classification tasks but possibly harms localization performance~\cite{Mahajan_2018_ECCV}. Li~\etal found that OpenImagesV4 pre-training~\cite{OpenImages} outperforms ImageNet pre-training when transferring to object detection and semantic segmentation~\cite{li2019detection_pretraining}. Shao~\etal showed similar findings using the Objects365 dataset~\cite{objects365}. This finding indicates supervised pre-training on a single classification task may not create representations general enough for many kinds of downstream applications.

Self-supervised training is a popular method for representation learning without supervised data~\cite{jing2020self,chen2020simple,grill2020bootstrap,he2019momentum,henaff2019data,tian2019contrastive}.
By forcing the representations of an image to agree with each other under data augmentation~\cite{becker1992self}, SimCLR and MoCo trained representations useful for downstream classification tasks~\cite{chen2020simple,he2019momentum}. Grill~\etal proposed the use of online and target neural networks for learning representations, which they evaluated on classification tasks as well as semantic segmentation, object detection, and depth estimation~\cite{grill2020bootstrap}. On the other hand, recent work has demonstrated the limitations of current self-supervised learning methods~\cite{purushwalkam2020demystifying}. They found that aggressive cropping, commonly used in self-supervised learning (such as those used in MoCo~\cite{he2019momentum}, PIRL~\cite{misra2020self}, SimCLR~\cite{chen2020simple} etc.), leads to representations that are occlusion invariant, which can be effective for downstream classification tasks. However, these representations are not necessarily invariant to other symmetries of natural images (such as viewpoint invariance), which might be necessary for other downstream tasks such as semantic segmentation~\cite{purushwalkam2020demystifying}.

\section{Method}

\subsection{Specialized Teacher Models}
We want to learn from a set of teachers that provide rich training signals with their pseudo labels. We adopt four teacher models including four important tasks in computer vision: classification, detection, segmentation, and depth estimation.
These tasks require visual understanding of objects and 3D geometry.
Examples of the pseudo labels can be found in Figure~\ref{fig:pseudo_labels}. We train the classification, detection, and segmentation teacher models from scratch on medium/large scale datasets (\eg, ImageNet\cite{ilsvrc}, Objects365~\cite{objects365}, COCO~\cite{Kirillov_2019_CVPR}).
For depth teacher model, we download the pre-trained checkpoint from the open-source repository~\cite{Ranftl2020}~\footnote{\textbf{\color{magenta}\url{https://github.com/intel-isl/MiDaS}}}.

\paragraph{Pseudo labeling: }
We transfer the knowledge in specialized teacher models to unlabeled or partially labeled datasets by pseudo labeling. We follow the practice in~\cite{zoph2020rethink} to generate pseudo labels for detection and segmentation. For detection,  we use a hard score threshold of 0.5 to generate pseudo box labels. For segmentation, we use a hard score threshold of 0.5 to generate semantic segmentation masks whereas pixels with a smaller prediction score are set to the \emph{ignore} label. For classification, we use soft labels, which contain the probability distribution of all classes, because we find the performance is better than hard labels. For depth, we simply use the predicted depth as pseudo labels without further processing.

\subsection{Multi-Task Student Model}
\paragraph{Model architecture: }
Our goal is to train the student with multiple tasks to learn general visual representations. The first thing to design is a model architecture that can share most of the parameters across tasks. We define three task categories: (1) classification, (2) object detection, (3) pixel-wise prediction. The pixel-wise prediction task includes semantic segmentation, depth estimation, and surface normal prediction. Each category of task shares the same feature representations in the backbone model.

We design the backbone model based on ResNet~\cite{resnet} and feature pyramid networks (FPN)~\cite{fpn}. Figure~\ref{fig:model} shows the overview of our architecture. We follow the common practice to design the feature representations for classification and detection tasks. We use \textit{C5} feature map (orange) for classification and \textit{\{P3, P4, P5, P6, P7\}} feature pyramid (cyan) for detection. We follow the practice in~\cite{zoph2020rethink} to fuse \textit{\{P3, P4, P5, P6, P7\}} into \textit{P2} feature map (green) for pixel-wise prediction. The fuse operation simply rescales all feature maps into level 2 and sums them (which does not introduce any new parameters).

Each task category shares the same head architecture. The classification head follows the ResNet design. It is a linear prediction layer followed by average pooled \textit{C5} features. The object detection task follows the head architecture in the Mask R-CNN~\cite{mrcnn}. We use 2 hidden convolution layers for RPN and 4 hidden convolution layers and 1 fully connected layers for Fast R-CNN. The pixel-wise prediction head has 3 convolution layers followed by the $C2$ features before the final linear prediction layer. If the student model learns from multiple tasks in the same task category (\eg, semantic segmentation and depth prediction), each task owns its task specific head without sharing their parameters.

\paragraph{Teacher-student training: }
We want to study the effectiveness of learning from pseudo labels. Therefore, we design the training of teacher and student models such that the main differences between them are in the dataset and the labels. Unlike model distillation~\cite{hinton2015distilling} and noisy student~\cite{xie2019self}, we use the same model capacity and data augmentation in both the teacher and the student training. Despite that a teacher can be trained with a more specialized architecture for its own task, we train the teacher and student models using the same architecture shown in Figure~\ref{fig:model}.

\begin{table*}[h]
\centering
\small
\begin{tabular}{lcc|lcc}
  \toprule
  \multicolumn{3}{c}{\textbf{Training Datasets}} &  \multicolumn{3}{|c}{\textbf{Evaluation Datasets}}\\
  \midrule
  Name & Task & Num Images & Name & Task & Num Images \\
  \midrule
  ImageNet~\cite{ilsvrc} & Classification & 1.2M & CIFAR-100~\cite{cifar} & Classification & 50k \\
  Objects365~\cite{objects365} & Detection & 600k & Pascal~\cite{everingham2010pascal} & Detection & 16.5k \\
  COCO~\cite{coco} & Segmentation & 118k & Pascal~\cite{everingham2010pascal} & Segmentation & 1.5k \\
  MiDaS~\cite{Ranftl2020} & Depth & 1.9M & NYU V2~\cite{silberman2012indoor} & Depth & 47k \\
  JFT~\cite{sun2017revisiting} & Classification & 300M & ADE~\cite{zhou2019semantic} & Segmentation & 20k \\
  & & & DIODE~\cite{vasiljevic2019diode} & Surface Normal & 17k \\
  \bottomrule
\end{tabular}
\caption{\textbf{Datasets using for MuST and for downstream fine-tuning evaluation}.  }
\vspace{-8px}
\label{tab:dataset}  
\end{table*}

\paragraph{Learning From Multiple Teachers: }
We propose Multi-Task Self-Training (MuST) to train a student model with multiple teachers. Prior multi-task learning works, which harnessed the information in multiple datasets, mainly focused on the scenario where each example is only labeled with one task or a few tasks~\cite{clark2019bam,Kokkinos_2017_CVPR}. In MuST, every image has supervision for all tasks. The labels may come from supervised or pseudo labels. For example, when training on ImageNet, we can use supervised labels for classification and pseudo labels for detection, segmentation, and depth.

Balancing the loss contribution for each task is an open research area in multi-task learning~\cite{kendall2017multi,chen2018gradnorm,chen2020just,yu2020gradient}. The loss of multi-task learning is the weighted sum of the losses from all tasks $L = \sum_iw_iL_i$. The weight $w_i$ decides the loss contribution for the task $i$.
In ImageNet experiments, we adopt $w_i = \frac{b_{s}lr_{it}}{b_{it}lr_{s}}$, where $b$ denotes the batch size, $lr$ denotes the learning rate, and the subscript denotes student or teacher model. The equation is derived from the scaling rule in~\cite{priya2017scaling}, which scales the learning rate propotionally with batch size. The only exception is the depth loss, of which we choose its weight by a parameter sweep. In our experiments on JFT-300M, we use the algorithm in~\cite{kendall2017multi} to learn $w_i$ for each task over the course of training.

\paragraph{Cross Dataset Training:}
MuST has the flexibility to leverage both labeled and unlabeled data. It can scale up the number of images by generating pseudo labels on the unlabeled data. Or it can use images which are partially labeled with one or more tasks. In our experiments, we show an example training across ImageNet, objects365, and COCO datasets. We use supervised labels whenever they are available and generate labels for all absent tasks using pseudo labels.

One challenge in cross dataset training is to balance the data coming from datasets of different sizes. Instead of designing sampling heuristics~\cite{clark2019bam}, we uniformly sample from the union of datasets. This works because every task is labeled on every image in MuST, thus we do not need to worry about under/over-sampling a task due to the imbalanced dataset size.

The second main difference compared to other self-training algorithm is that the supervised and pseudo labels are treated equally. We do not batch the examples of supervised and pseudo labels independently and assign them different weights like in~\cite{zoph2020rethink,xie2019self}. The images are uniformly sampled from the union of datasets and put into one mini-batch. Each example shares the same weight on its loss regardless if the loss is computed against a supervised or a pseudo label. This makes MuST significantly simpler to use and to scale with multiple tasks.

\subsection{Transfer Learning}
To evaluate the representational quality of MuST and other baseline representations, we fine-tune them on a suite of downstream computer visions tasks. We adopt \textit{end-to-end} fine-tuning instead of linear probe to the performance of each fine-tuning task. We fine-tune on CIFAR-100 classification, Pascal detection, Pascal semantic segmentation, NYU depth, ADE semantic segmentation and DIODE surface normal. Also note that all downstream datasets are different than the ones the specialized teacher models are trained on. Furthermore, surface normal prediction is a task that no specialized teacher model was trained for, testing the robustness of representations to held out tasks.

When fine-tuning a representation on a downstream task we sweep over the learning rate and number of training steps (See Appendix for full details). This allows for fair comparison between different representations.

\section{Experiments}
\subsection{Experimental Settings}

\paragraph{Training Datasets:}
Table~\ref{tab:dataset} provides an overview of the datasets we use in the experiments.
We experiment with four different datasets and tasks for training our supervised teacher models. These supervised models will then be the ones to generate pseudo labels on unlabeled/partially labeled images.

\paragraph{Evaluation Datasets:}
Next we describe the datasets that all of our representations will be fine-tuned on. Table~\ref{tab:dataset} provides the list. We have different datasets with a total of five different tasks. Note the Surface Normal task is never used as a training task to test the task generality of the representations.

\subsection{Learning with Multi-Task Self-Training}
We run  experiments to compare our MuST representation learning algorithm to state-of-the-art self-supervised and supervised learning methods.

\begin{table*}[h!]
\centering
\small
\begin{tabular}{lc|cccccc}
  \toprule
  \multicolumn{2}{c}{\textbf{Settings}} &  \multicolumn{6}{|c}{\textbf{Transfer Learning Performance}}\\
  \midrule
  Method & Epochs & CIFAR-100 & Pascal & Pascal & NYU & ADE & DIODE  \\
  & & Classification & Detection & Segm. & Depth & Segm. & Normal  \\
  \midrule
  Self-supervised (SimCLR~\cite{chen2020simple}) & 800 & \textbf{87.1} & 83.3 & 72.2 & 83.7 & 41.0 & \textbf{52.8} \\
  \midrule
  ImageNet Supervised & 90 & 85.4 & 79.3 & 70.6 & 81.0 & 39.8 & 48.9\\
  + Multi-task Pseudo Labels
& 90  & 86.3 \plus{0.9} & \textbf{85.1} \plus{5.8} & \textbf{80.6} \plus{10.0} & \textbf{87.8} \plus{6.8} & \textbf{43.5} \plus{3.7} & \textbf{52.7} \plus{3.8} \\
  \bottomrule
\end{tabular}
\caption{\textbf{Multi-Task Self-Training (MuST) outperforms supervised and self-supervised representations on ImageNet}. We compare MuST to state-of-the-art self-supervised and supervised learning using the same pre-training dataset (ImageNet). MuST learns more general features and achieves the best performance on $4 / 6$ downstream fine-tuning tasks. The performance differences show the impact of different training objectives. }
\label{tab:improve_imagenet}  
\end{table*}

\paragraph{MuST Improves Pre-training on ImageNet:}
Table~\ref{tab:improve_imagenet} compares the MuST algorithm to self-supervised and supervised learning on ImageNet. On a suite of 6 downstream tasks MuST representations improves over state-of-the-art self-supervised learning and supervised learning on $4$ and $5$ tasks, respectively. MuST makes use of not only the ImageNet classification labels, but also pseudo labels generated from supervised models trained on Objects365 detection, COCO semantic segmentation, and MiDas depth. This additional information being trained on for ImageNet images leads to much more generalizable feature representations. We observe that self-supervised and supervised pre-training on ImageNet does not learn features that generalize nearly as well to tasks other than image classification.

\begin{table*}[h!]
\centering
\small
\begin{tabular}{l|cccccc}
  \toprule
  \multicolumn{1}{c}{\textbf{Settings}} &  \multicolumn{6}{|c}{\textbf{Transfer Learning Performance}}\\
  \midrule
  Method  & CIFAR-100 & Pascal & Pascal & NYU & ADE & DIODE  \\
  & Classification & Detection & Segm. & Depth & Segm. & Normal  \\
  \midrule
  ImageNet Supervised & 85.4 & 79.3 & 70.6 & 81.0 & 39.8 & 48.9 \\
  + Depth Pseudo Labels & 84.4\minus{1.0} & 79.3\plus{0.0} & 71.0\plus{0.4} & 86.0\plus{5.0} & 39.5\minus{0.3} & 51.3\plus{2.4} \\
  + Depth / Segm. Pseudo Labels & 85.3\minus{0.1} & 81.6\plus{2.3} & 78.6\plus{8.0} & 87.2\plus{6.2} & 41.5\plus{1.7} & 52.4\plus{3.5} \\
  + Depth / Segm. / Detection Pseudo Labels & \textbf{86.3}\plus{0.9} & \textbf{85.1}\plus{5.8} & \textbf{80.6}\plus{10.0} & \textbf{87.8}\plus{6.8} & \textbf{43.5}\plus{3.7}  & \textbf{52.7}\plus{3.8}\\
  \bottomrule
\end{tabular}
\caption{\textbf{Multi-Task Self-Training (MuST) benefits from increasing the number of different pseudo label tasks.} We add depth, segmentation, and detection pseudo labels in addition to supervised ImageNet classification labels and test the representational quality. The results reveal that adding pseudo labels from more tasks leads to more general pre-trained models. All models are trained for 90 epochs on ImageNet.}
\label{tab:inrease_tasks}  
\end{table*}

\paragraph{MuST Improves With More Tasks/Datasets For Learning General Features:}
The MuST algorithm makes use of pseudo labels generated from independent supervised models trained on different datasets. We next study the importance of having pseudo labels being generated from multiple different teacher models trained on different tasks/datasets. Table~\ref{tab:inrease_tasks} shows the representational quality improvement starting from using only supervised ImageNet labels and then adding three different types of pseudo labels obtained from three different datasets. As we continue to add pseudo labels from different tasks/datasets our representations improve in quality. For each new task added we obtain strong improvement across all 6 of our downstream tasks.

\paragraph{Pseudo Labels Preserve Transfer Learning Performance of Teacher Model:}
We next study how effectively pseudo labels preserve the transfer learning performance of teacher models trained on supervised datasets. To test this we train two supervised teacher models: object detection model on Objects365 and semantic segmentation model on COCO. The first two rows in Table~\ref{tab:teacher_student} shows their supervised learning performance and their transfer learning performance on 6 downstream tasks. Next we generate pseudo labels on two datasets without labels: ImageNet (1.2M images) and JFT (300M images) . Now we train models from scratch on the pseudo labels on ImageNet and JFT. The next 4 rows in Table~\ref{tab:teacher_student} reveal these results. We observe for both object detection and segmentation pseudo labels we obtain a degradation in the supervised learning quality (e.g. 26.1 vs 20.6/20.7), but that when the representations are \emph{transferred} they obtain similar or better transfer learning performance than the teacher model. Furthermore, the representations obtained by training on JFT over ImageNet typically lead to better transfer learning performance, which reveals the scalability of the MuST method. As we get more and more unlabeled data, our method can easily take advantage of it and the representational quality improves.

\begin{table*}[h!]
\centering
\small
\begin{tabular}{lc|cc|cccccc}
  \toprule
  \multicolumn{2}{c|}{\textbf{Settings}} &  \multicolumn{2}{c|}{\textbf{Performance}} & \multicolumn{6}{c}{\textbf{Transfer
  Learning Performance}}\\
  \midrule
  Task  & Train Dataset & Obj365 & COCO & CIFAR-100 & Pascal & Pascal & NYU & ADE & DIODE  \\
  & & Detection & Segm. & Classification & Detection & Segm. & Depth & Segm. & Normal \\ 
  \midrule
  & & \multicolumn{2}{c|}{\textbf{Teacher Model}} \\
  \midrule
  Detection & Objects365 & 26.1 & --- & 84.0 & 87.6 & 78.8 & 90.1 & 46.0 & 55.6 \\
  \rowcolor{gray!15}
  Segmentation & COCO & --- & 53.8 & 80.8 & 82.2 & 80.2 & 86.6 & 42.8 & 51.0 \\
  \midrule
  & & \multicolumn{2}{c|}{\textbf{Student Model}} \\
  \midrule
  Detection & ImageNet & 20.6 & --- & 83.2 & 86.0 & 78.5 & 88.5 & 44.7 & 55.2\\
  Detection & JFT & 20.7 & --- & 85.2 & 87.7 & 79.5 & 89.6 & 45.4 & 55.0 \\
  \rowcolor{gray!15}
  Segmentation & ImageNet & --- & 55.5 & 82.3 & 80.5 & 79.2 & 86.3 & 41.8 & 51.2  \\
  \rowcolor{gray!15}
  Segmentation & JFT & --- & 49.0 & 83.1 & 82.8 & 78.2 & 86.6 & 41.9 & 51.6 \\
  \bottomrule
\end{tabular}
\caption{\textbf{Models trained on supervised data or pseudo labeled data have similar transfer learning performance.} Results comparing how representations transfer if they are trained on supervised data or on pseudo labels that are generated by the supervised model. Pseudo labels effectively compress the knowledge in a supervised dataset. The performance of student models increases with the size of the unlabeled dataset. As the unlabeled dataset size increased, the performance of student model increases. This reveals the scalability of MuST. All student models are trained for the same training iterations (90 ImageNet epochs and 0.36 JFT epochs).}
\label{tab:teacher_student}  
\end{table*}

\paragraph{Multi-Task Self-Training Across Datasets:}
MuST utilized pseudo labels generated from teacher models trained on different supervised learning datasets. A natural comparison is then to see how 
MuST compared against supervised multi-task supervised training where a model is trained on the union of the datasets and labels~\cite{Kokkinos_2017_CVPR}.
Table~\ref{tab:multi_task} compares the representational quality of MuST versus supervised multi-task training on three datasets: ImageNet, COCO and Objects365. For multi-task training we sample examples from the datasets with equal probability. Sampling examples with probabilities proportional to the size of the datasets does not work well. Because ImageNet and Objects365 datasets are much larger than COCO dataset and as a result for a batch size of 256 only 15 examples have non zero loss values for segmentation. On the other hand, for MuST \emph{every image has any type of label} and we can sample examples with probabilities proportional to the size of datasets.
When comparing the representation qualities MuST obtains the best performance on $6 / 6$ downstream tasks.

\begin{table*}[h!]
\centering
\small
\begin{tabular}{l|cccccc}
  \toprule
  \multicolumn{1}{c}{\textbf{Settings}} & \multicolumn{6}{|c}{\textbf{Transfer Learning Performance}}\\
  \midrule
  Method  & CIFAR-100 & Pascal & Pascal & NYU & ADE & DIODE  \\
  & Classification & Detection & Segm. & Depth & Segm. & Normal \\ 
  \midrule
  Supervised Multi-Task & 85.3 & 85.1 & 82.1 & 87.6 & 43.9 & 53.4 \\
  Supervised Multi-Task + Pseudo Labels & \textbf{86.3} \plus{1.1} & \textbf{86.2} \plus{1.1} & \textbf{82.3} \plus{0.2} & \textbf{88.2} \plus{0.6} & \textbf{45.4} \plus{1.5} & \textbf{54.7} \plus{1.3} \\
  \bottomrule
\end{tabular}
\caption{\textbf{Comparing Multi-Task Training versus Multi-Task Self-Training.} We compare MuST against a baseline of doing supervised multi-task training on the union of all teacher datasets. We use three datasets: ImageNet, COCO and Objects365. Supervised model is jointly trained on the supervised labels of these three datasets. MuST trains jointly on all three supervised and pseudo labels generated by the teacher models. The transfer learning performance gets strong improvements by incorporating pseudo labels into every image.}
\label{tab:multi_task}  
\end{table*}

\begin{table*}[h!]
\centering
\small
\begin{tabular}{lc|cccccc}
  \toprule
  \multicolumn{2}{c}{\textbf{Settings}} &  \multicolumn{6}{|c}{\textbf{Transfer Learning Performance}}\\
  \midrule
  Method & Epochs & CIFAR-100 & Pascal & Pascal & NYU & ADE & DIODE  \\
  & & Classification & Detection & Segm. & Depth & Segm. & Normal  \\
  \midrule
  Self-Supervised with JFT images (SimCLR~\cite{chen2020simple}) & 1 & 85.6 & 82.4 & 71.0 & 83.7 & 41.4 & 54.4\\
  Self-Supervised with JFT images (SimCLR~\cite{chen2020simple}) & 2 & 85.8 & 83.7 & 73.3 & 84.3 & 42.2 & 55.3\\
   Self-Supervised with JFT images (SimCLR~\cite{chen2020simple}) & 5 & 86.1 & 84.1 & 74.9 & 84.8 & 43.0 & 56.0\\
  \midrule
  JFT supervised & 3 & 87.7 & 84.6 & 78.2 & 86.0 & 43.4 & 50.7\\
  JFT supervised & 5 & 88.6 & 84.9 & 79.7 & 86.1 & 44.3 & 51.1\\
  JFT supervised & 10 & \textbf{89.6} & 85.2 & 80.4 & 86.5 & 45.7 & 53.1 \\
  \midrule
  Multi-Task Pseudo Labels & 2.5 & 87.6 & 87.8 & 82.2 & \textbf{89.8} & 47.0 & 56.2 \\
  JFT supervised + Multi-Task Pseudo Labels & 2 & 88.3\plus{0.5} & \textbf{87.9}\plus{0.1} & \textbf{82.9}\plus{0.7} & 89.5\minus{0.3} & \textbf{47.2}\plus{0.2} & \textbf{56.4}\plus{0.2}\\
  \bottomrule
\end{tabular}
\caption{\textbf{Scaling Multi-Task Self-Training to 300M images.} We repeat the experiments in Table~\ref{tab:improve_imagenet} on the JFT dataset (300M images with classification labels). The supervised learning benefits more from the additional images and annotations compared to the self-supervised SimCLR algorithm.}
\label{tab:jft_r152}  
\end{table*}

\subsection{Scaling Multi-Task Self-Training}
One benefit of MuST is that it can scale to unbounded amounts of unlabeled images. To test this hypothesis we move from the ImageNet setup with 1.2M images to JFT with 300M images.

\begin{figure}[t!]%
\centering
\begin{tabular}{ccc}%
\includegraphics[width=0.3\linewidth]{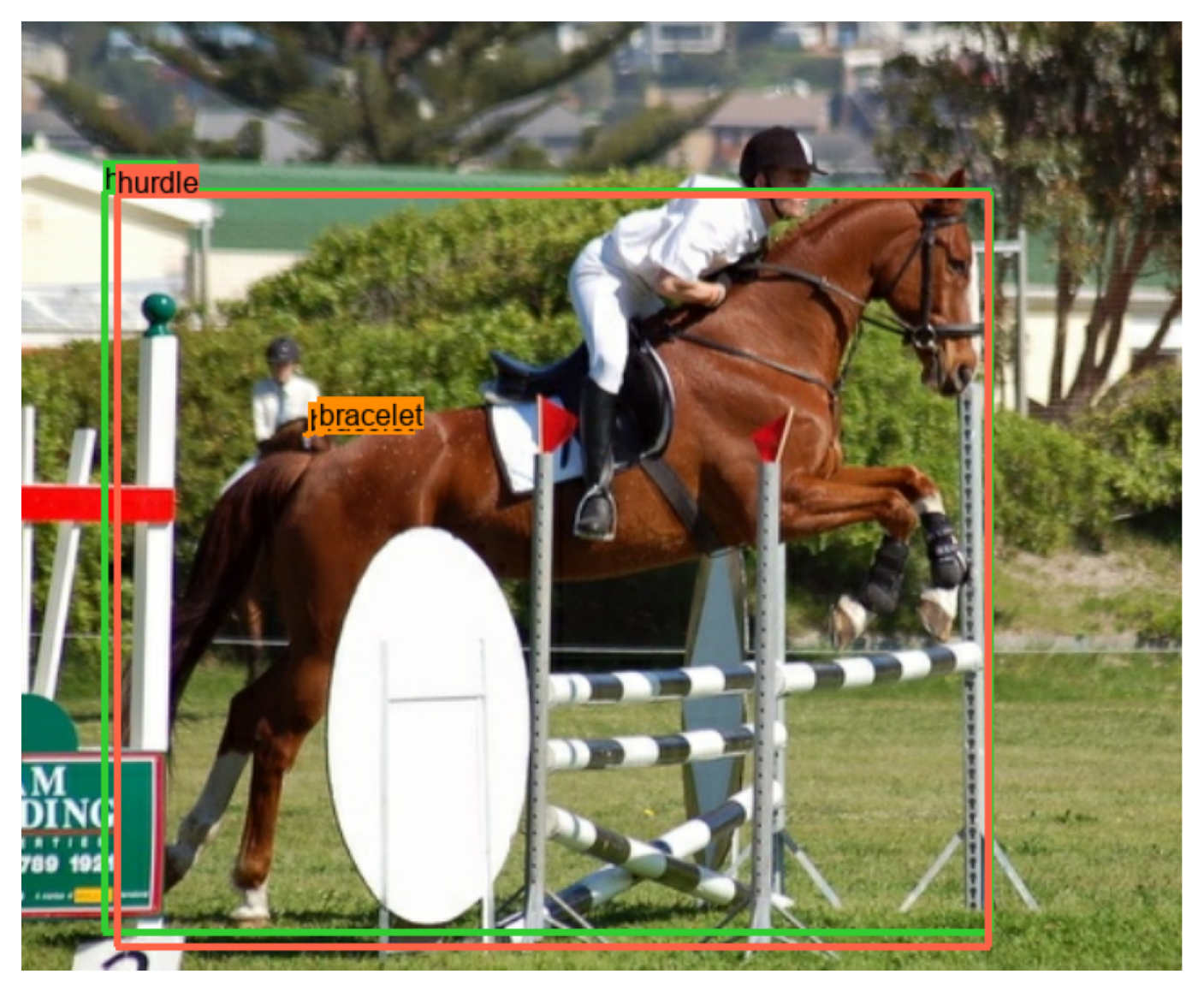}&%
\includegraphics[width=0.3\linewidth]{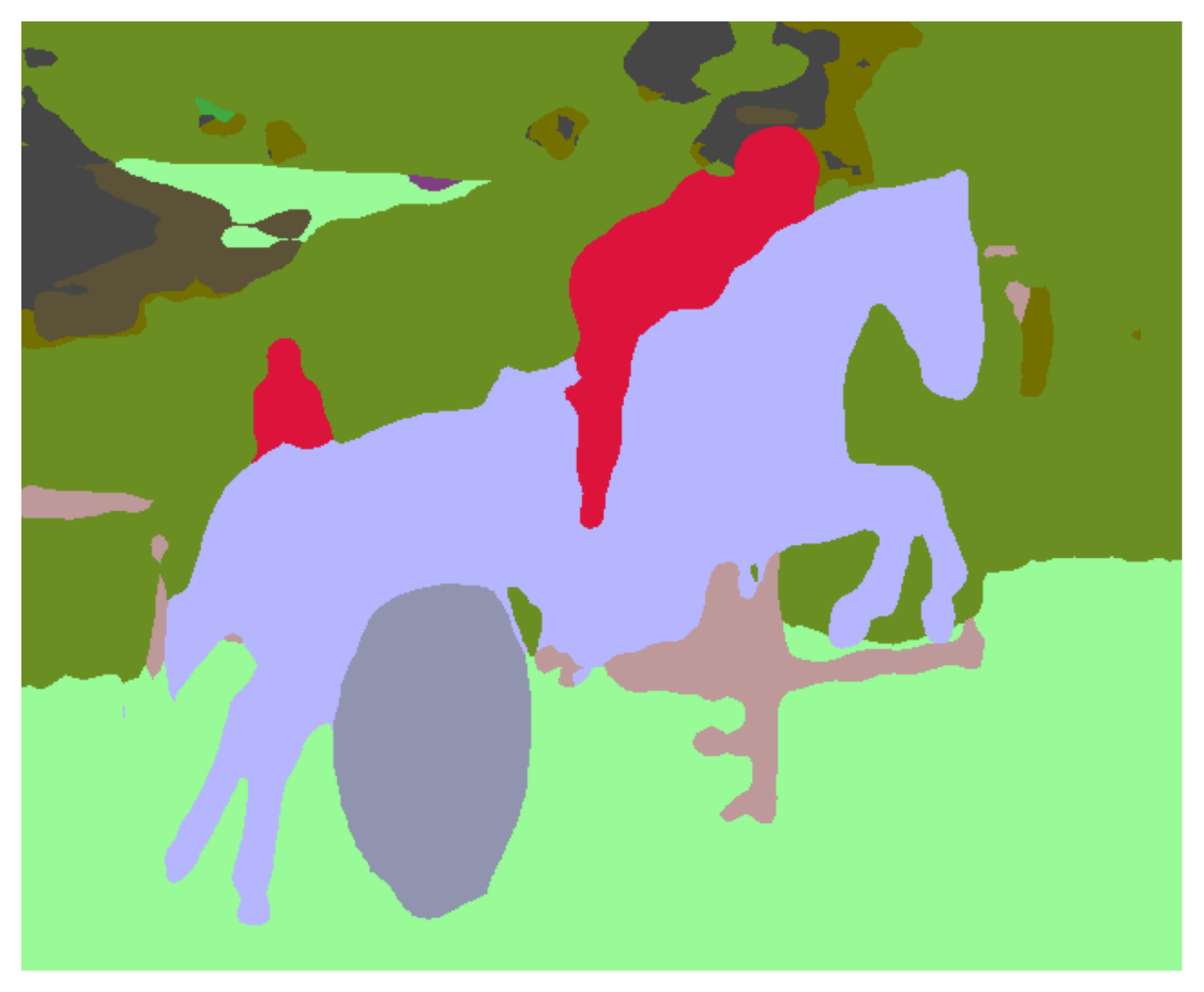}&%
\includegraphics[width=0.3\linewidth]{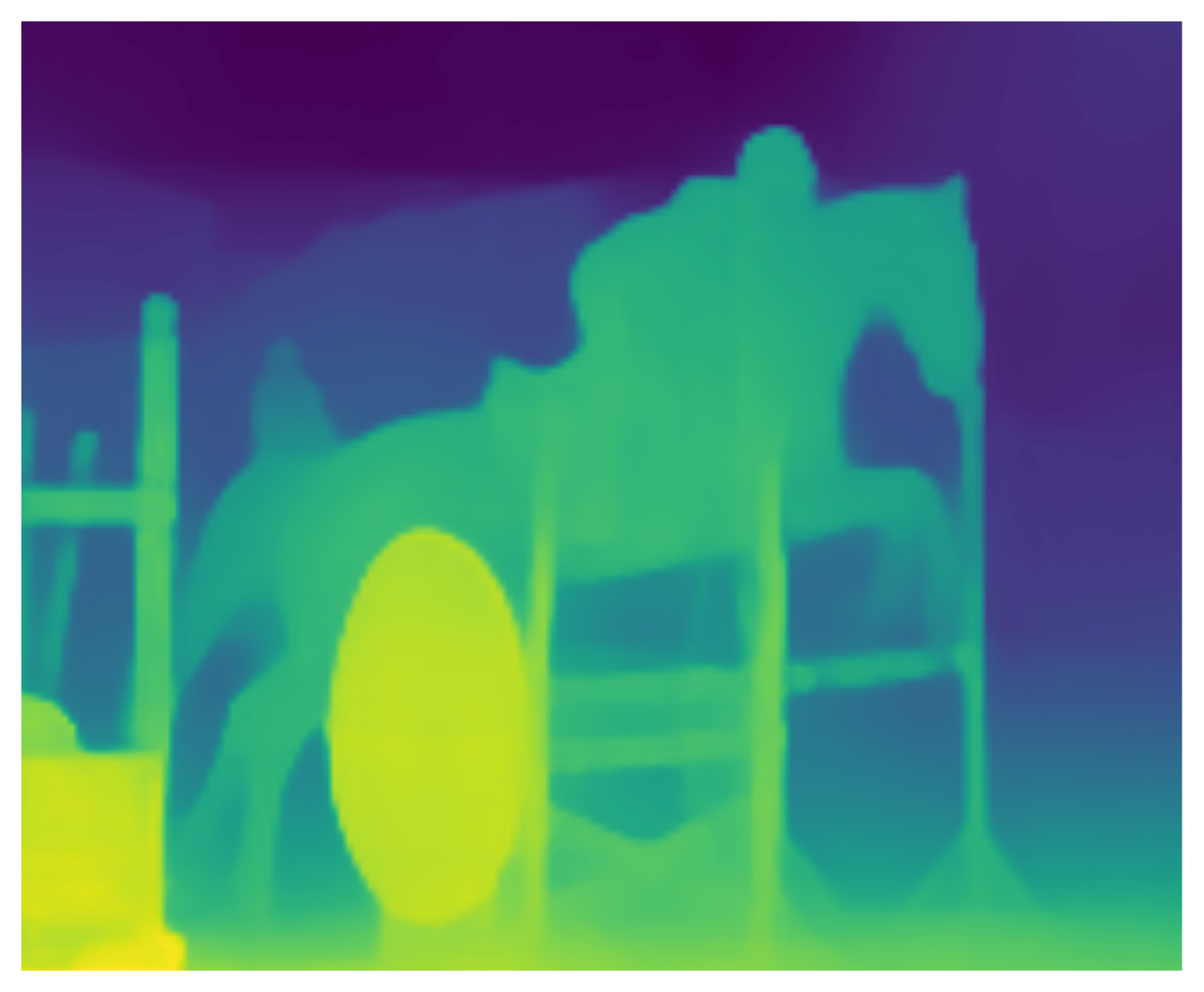}\\%
\end{tabular}
  \caption{\textbf{Visualization of the predictions generated by a multi-task student model.} The MuST student model not only learns general feature representations, but also makes high quality visual predictions with a single model.
  }%
  \label{fig:student_model_predictions}%
\end{figure}

\paragraph{Scaling Dataset Size and Training Iterations: }
Now instead of generating pseudo labels on 1.2M images, we scale the MuST training to have all three supervised teacher models to generate pseudo labels on 300M images. This process is trivially parallelizable, which makes the overall runtime low compared to training of the models. Table~\ref{tab:jft_r152} shows the comparison of MuST vs self-supervised learning and supervised learning on the JFT dataset. On $5 / 6$ downstream tasks MuST outperforms the self-supervised SimCLR algorithm when using the same unlabeled data. We also train a supervised baseline on the multi-class labels available on JFT and find that MuST, using only the unlabeled images, outperforms the representation on $5 / 6$ downstream tasks. This is quite impressive considering the total sum of supervised images that MuST indirectly makes use of from the pseudo labels is only about 3.7M images compared to the 300M labeled JFT images. Adding JFT supervised labels can further improve the performance on image classification and segmentation, showing the flexibility of MuST in using labeled and unlabeled data. Lastly, the student model not only learns general features for transferring, it is also capable of generating high quality predictions for multiple tasks. Figure~\ref{fig:student_model_predictions} shows the predictions made by our strongest model.

\begin{figure}[t!]
    \centering
    \includegraphics[width=0.45\textwidth]{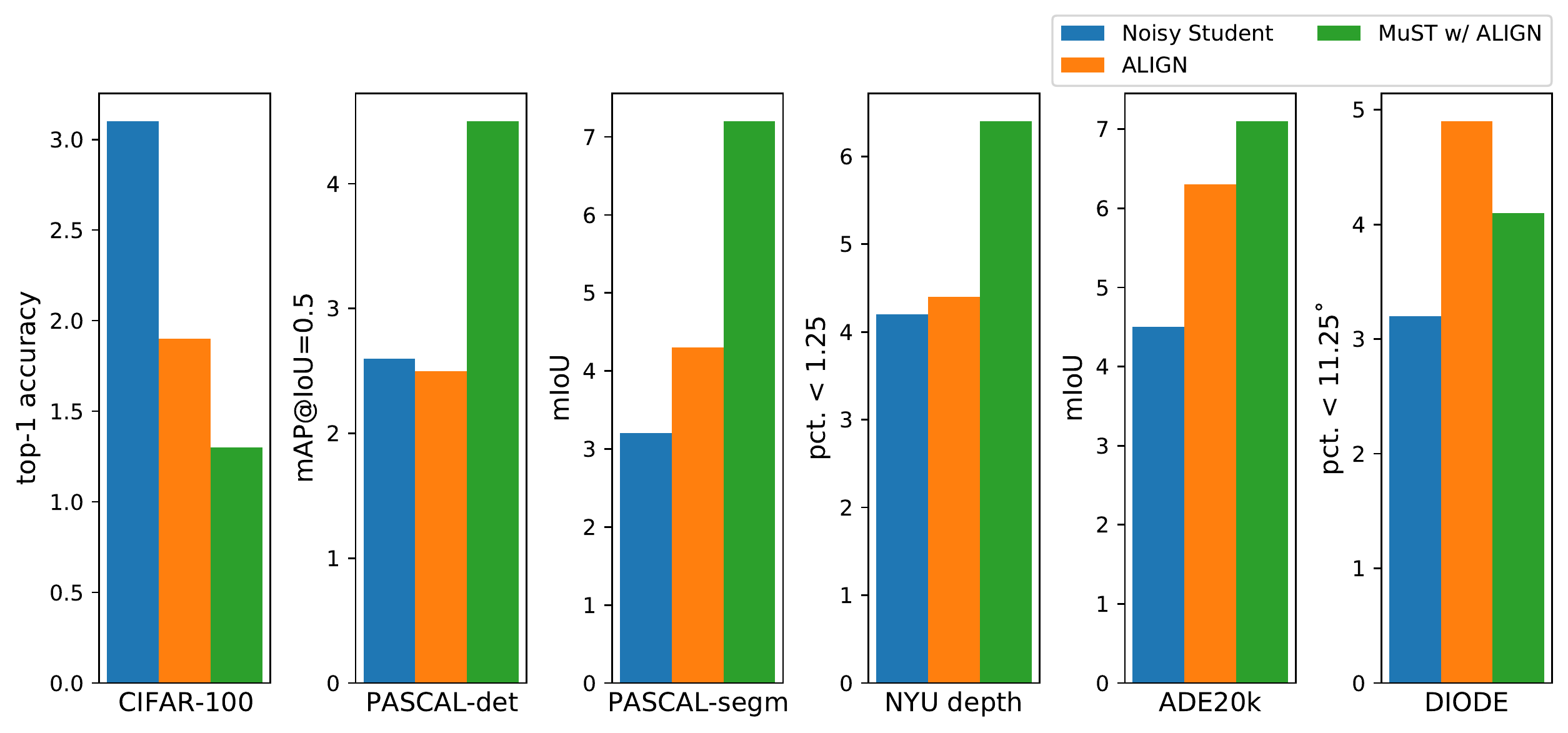}
    \caption{\textbf{Relative transfer learning performance gains over the ImageNet pre-trained model~\cite{tan19enet}}. Checkpoints trained with more data or labels typically provide gains on transfer learning to downstream tasks. Fine-tuning the EfficientNet-B7 ALIGN checkpoint with MuST can further improve transfer learning performance for 4/6 downstream tasks.}
    \label{fig:relative_gain}
\end{figure}

\paragraph{Bootstrapping from Pre-trained Models.}
Next we study if MuST can improve upon checkpoints trained with billions of training examples. We use ALIGN checkpoints~\cite{chao2021align}, which are trained with 1.8B image-text pairs, to initialize parameters for training both teacher models and the student model. We use the same teacher model tasks as our previous experiments. The pseudo labels are generated on JFT-300M dataset and the MuST student model is trained on JFT for 1 epoch. Figure~\ref{fig:relative_gain} shows relative transfer learning performance gains of Noisy Student~\cite{xie2019self}, ALIGN~\cite{chao2021align}, and MuST w/ ALIGN compared to the ImageNet checkpoint~\cite{tan19enet}. The figure shows MuST w/ ALIGN improves the ALIGN checkpoint by respectable margins for 4 out of the 6 downstream tasks. The performances are slightly worse for CIFAR-100 and DIODE surface normal prediction. We repeat the experiments with EfficientNet-L2 architecture and train the student model for 0.36 epoch on JFT. We report 4 downstream tasks showing improvements over the ALIGN checkpoint in Table~\ref{tab:sota_with_l2}. We find the student model trained with MuST improves the large ALIGN EfficientNet-L2 checkpoint and is competitive to the state-of-the-art models specialized for each dataset and task. Notably, MuST provides checkpoints ready to be fine-tuned for short iterations to achieve state-of-the-art performance while typical self-training methods~\cite{zoph2020rethink} require pseudo labeling and long training iterations for each downstream task.
\begin{table}[t!]
\centering
\small
\begin{tabular}{l|cccc}
  \toprule
  \multicolumn{1}{c}{\textbf{Settings}} &  \multicolumn{4}{|c}{\textbf{Transfer Learning Performance}}\\
  \midrule
  Method & Pascal & Pascal & NYU & ADE \\
   & Detection & Segm. & Depth & Segm. \\
  \midrule
  Previous SoTA & \textbf{89.3}~\cite{ghiasi2021copypaste} & \textbf{90.0}~\cite{zoph2020rethink} & 90.4~\cite{ranftl2021dpt} & \textbf{54.1}~\cite{cheng2021maskformer} \\
  ALIGN~\cite{chao2021align} & 86.2 & 86.6 & 91.1 & 54.0 \\
  MuST w/~\cite{chao2021align} & 88.2 & \textbf{89.8} & \textbf{91.9} & \textbf{54.3} \\
  \bottomrule
\end{tabular}
\caption{\textbf{MuST checkpoints are versatile and achieve competitive performance compared to state-of-the-art models.} MuST improves the transfer learning performance of the ALIGN EfficientNet-L2 checkpoint on these four downstream tasks.}
\label{tab:sota_with_l2}
\end{table}

\section{Discussion}

\paragraph{Which pre-training method performs the best with large scaling training?} Although self-supervised learning can outperform supervised learning on the ImageNet size dataset (1.3 million images/1k classes), supervised learning is still a better pre-training method on JFT size dataset (300 million images/18k classes). The gap may be compensated by training with more unlabeled data for self-supervised learning. However, self-training can also expand one or multiple supervised models by generating pseudo labels on unlabeled data. Overall, both self-supervised and self-training are able to scale, but at the moment self-training presents better performance in learning general features. A promising direction is to combine self-supervised and self-training for representation learning~\cite{zhang2020pushing,du2020selftraining,xu2020selftraining}.

\paragraph{Why use MuST over self-supervised learning?} Both of the methods are scalable with the unlabeled training data, however, MuST can easily combine together all labeled and unlabeled data. 
However, self-supervised learning relies on the generalization from the pre-text task to downstream tasks, which does not always give good performance.
It is easier to design pseudo labels if the downstream tasks of interest are known in advance. MuST also generalizes to unseen tasks (\eg, surface normal prediction) given the tasks of the teacher model in this paper.

\section{Conclusion}
In this paper, we present MuST, a scalable multi-task self-training method for learning general representations. We compare with supervised and self-supervised learning approaches on ImageNet and JFT and evaluate on 6 datasets including visual recognition, localization, and 3D geometry prediction. We show that MuST outperforms or is on par with supervised and self-supervised learning on 5 out of 6 transfer learning tasks, except the classification task.
Moreover, MuST can improve upon already strong checkpoints trained with billions of examples.
The results show multi-task self-training is a scalable pre-training method and is able to learn general feature representations. We hope this work will encourage further research towards creating universal visual representations.

\section*{Acknowledgements}
We would like to thank Yin Cui, Aravind Srinivas, Simon Kornblith, and Ting Chen for valuable feedback.

\appendix
\begin{appendix}

\section{Details of Training and Evaluation Datasets}

\subsection{Training Datasets}
In this section, we describe 5 datasets we use to train teacher models.

\paragraph{ImageNet:} ImageNet~\cite{ilsvrc} is a classification dataset with 1.2M training images and 1000 unique classes. All of its images are center cropped and have one primary object per image.

\paragraph{Objects365:} Objects365~\cite{objects365} is an object detection dataset that has 365 different classes and 600k training images.

\paragraph{COCO:} The COCO dataset~\cite{coco} contains 118k images that has a variety of different labels (e.g. object detection, instance segmentation, panoptic segmentation). For all experiments we use its pantopic segmentation labels. 

\paragraph{MiDaS:} The MiDaS depth model~\cite{Ranftl2020} that is used for generating our depth pseudo labels is trained on a diverse set of 5 depth datasets. The 5 depth datasets are DIML Indoor~\cite{kim2018deep} (220k images), MegaDepth~\cite{li2018megadepth} (130k images), ReDWeb~\cite{xian2018monocular} (3600), WSVD~\cite{wang2019web} (1.5M), and 3D movies (75k). The model is trained to be invariant to the depth range and scale across all datasets, leading to a model that generates robust pseudo labels.

\paragraph{JFT:} JFT~\cite{sun2017revisiting} is a large-scale image multi-label classification dataset with 300M labeled images. This dataset is used to test the scale of MuST and various self-supervised learning algorithms.

\subsection{Evaluation Datasets}
Next we describe the datasets that all of our representations will be fine-tuned on. We have different datasets with a total of five different tasks. Note the Surface Normal task is never used as a training task to test the task generality of the representations.

\paragraph{CIFAR-100:} CIFAR-100 is a classification dataset with 50k images and 100 unique classes.
\paragraph{PASCAL Detection:} The Pascal Detection dataset~\cite{everingham2010pascal} is an object detection dataset with 20 unique classes. We train the models on the \texttt{trainval} sets of PASCAL VOC 2007 and PASCAL VOC 2012 which include 16.5k images.
\paragraph{PASCAL Segmentation:} The Pascal Segmentation dataset~\cite{everingham2010pascal} is a semantic segmentation dataset with 20 unique classes.  We train the models on the \texttt{train} set of the PASCAL VOC 2012 segmentation dataset which has 1.5k images.
\paragraph{NYU Depth V2:} The NYU Depth v2 dataset~\cite{silberman2012indoor} is a depth estimation dataset that contains 47584 train images and 654 validation images. 
\paragraph{ADE Segmentation:} ADE20k~\cite{zhou2019semantic} is a segmentation dataset that contains 20k images with 150 object and stuff classes. The dataset contains a wide variety of different indoor and outdoor scenes along with object classes.
\paragraph{DIODE Surface Normal:} The DIODE dataset~\cite{vasiljevic2019diode} is a depth and surface normal dataset that contains 16884 images. The dataset contains a diverse set of both indoor and outdoor scenes for training and testing. We only make use of the surface normal labels.

\section{Implementation Details}
\subsection{Training Teacher Models}
In this section, we introduce the details of training teacher models, which are used to generate pseudo labels in MuST. All the models are trained with a ResNet-152 backbone model.

\paragraph{Objects365 Detection:}
We use batch size 256 and train for 140 epochs. The image size is $640\time640$. We apply scale jittering [0.5, 2.0] (i.e. randomly resample image between $320\times320$ to $1280\times1280$ and crop it to $512\times512$). The learning rate is 0.32 and the weight decay is set as 4e-5. The model is trained from random initialization. The final performance is 26.1 AP.

\paragraph{COCO Segmentation:}
We use the annotations in COCO panoptic segmentation dataset~\cite{Kirillov_2019_CVPR}. We train a semantic segmentation model that only predicts the semantic class for each pixel, instead of also predicting the object instance. We use batch size 256 and train for 384 epochs. The image size is $896\time896$. We apply scale jittering [0.5, 2.0]. The learning rate is 0.32 and the weight decay is set as 4e-5. The model is trained from random initialization. The final performance is 53.8 mIoU.

\paragraph{MiDaS Depth:}
We directly download the pre-trained MiDaS from the github repository and use it as a teacher model to generate pseudo labels.

\paragraph{ImageNet Classification:}
We use batch size 2048 and train for 400 epochs. The image size is $224\time224$. The learning rate is 0.8 and weight decay is 4e-5. We apply random augmentation~\cite{cubuk2019randaugment} (2L-15M, 2 layers with magnitude 15) and label smoothing (0.1) to regularize the model training. The final performance is 81.6 top-1 accuracy.

\subsection{Training Multi-Task Student Models}
We use a batch size 256 for training student models in our experiments. The image size is $640\time640$. We apply scale jittering [0.5, 2.0] during training. The weight decay is 4e-5 in ImageNet experiments and 3e-6 in JFT experiments. No random augmentation~\cite{cubuk2019randaugment} or label smoothing is applied.

\subsection{Fine-tuning on Evaluation Datasets}
For fine-tuning we initialize the parameters in the ResNet and FPN backbone with a pre-trained model and randomly initialize the rest of the layers. 
We perform \textit{end-to-end} fine-tuning with an extensive grid search of the combinations of learning rate and training steps to ensure each pre-trained model achieves its best fine-tuning performance. 
We experiment with different weight decays but do not find it making a big difference and set it to 1e-4. All models are trained with cosine learning rate for simplicity. 
Below we describe the dataset, evaluation metric, model architecture, and training parameters for each task.

\paragraph{CIFAR-100:} We use standard CIFAR-100 train and test sets and report the top-1 accuracy. We resize the image resolution to $256\times256$. We replace the classification head in the pre-trained model with a randomly initialized linear layer that predicts 101 classes, including background. We use a batch size of 512 and search the combination of training steps from 5000 to 20000 and learning rates from 0.005 to 0.32. We find the best learning rate for SimCLR (0.16) is much higher than the supervised model (0.005). This trend holds for the following tasks.

\paragraph{PASCAL Segmentation:} We use PASCAL VOC 2012 train and validation sets and report the mIoU metric. The training images are re-sampled into $512\times512$ with scale jittering [0.5, 2.0]. We initialize the model from the pre-trained backbone and FPN~\cite{fpn} layers. We remove the pre-trained segmentation head and train from a randomly initialized head. We use a batch size of 64 and search the combination of training steps from 5000 to 20000 and learning rates from 0.005 to 0.32.

\paragraph{PASCAL Detection:}
We use PASCAL VOC 2007+2012 trainval set and VOC 2007 test set and report the $AP_{50}$ with 11 recall points to compute average precision. The training images are resampled into $896\time896$ with scale jittering [0.5, 2.0]. we initialize the model from the pre-trained backbone and FPN~\cite{fpn} layers and randomly initialize the heads.
We use a batch size of 32 and search the combination of training steps from 5000 to  20000 and learning rates from 0.005 to 0.32.

\paragraph{NYU Depth:} We use NYU depth v2 dataset with 47584 train and 654 validation images. We report the percentage of predicted depth values within 1.25 relative ratio compared to the ground truth. The training images are resampled into $640\time640$ with scale jittering [0.5, 2.0]. we initialize the model from the pre-trained backbone and FPN~\cite{fpn} layers and randomly initialize the heads. We use a batch size of 64 and search the combination of training steps from 10000 to 40000 and learning rates from 0.005 to 0.32.

\paragraph{DIODE:} We use DIODE outdoor dataset with 16884 train and 446 validation images. We report the percentage of the angle error less than 11.25\degree. We use the original image resolution $768\time1024$ for training and evaluation. The training image is applied with scale jittering [0.5, 2.0]. we initialize the model from the pre-trained backbone and FPN~\cite{fpn} layers and randomly initialize the heads. We use a batch size of 32 and search the combination of training steps from 20000 to 80000 and learning rates from 0.01 to 0.16.

\section{Visualization of Student Model Predictions}
\begin{figure*}[ht!]%
\centering%
\begin{tabular}{ccc}%
\includegraphics[width=0.3\linewidth]{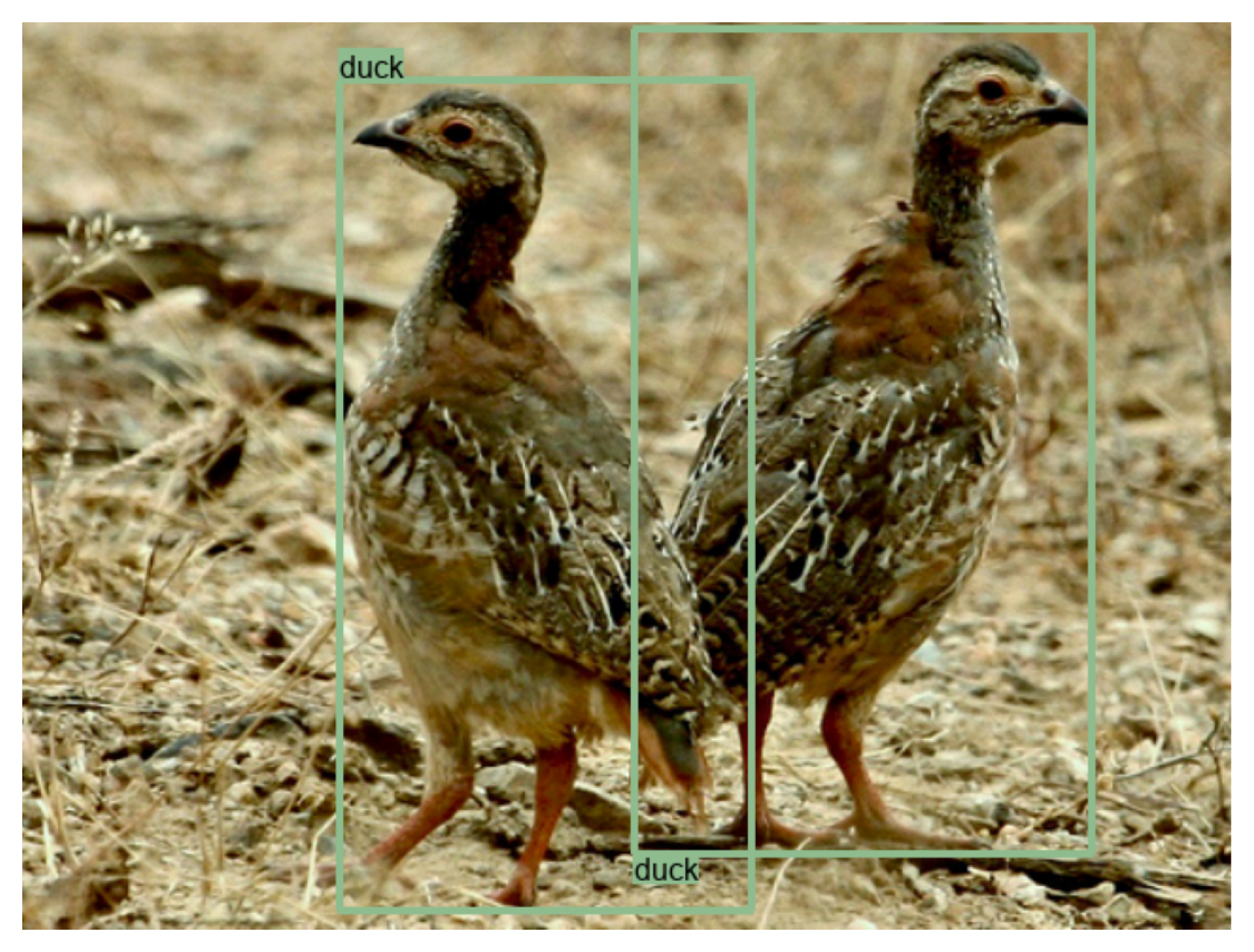}&%
\includegraphics[width=0.3\linewidth]{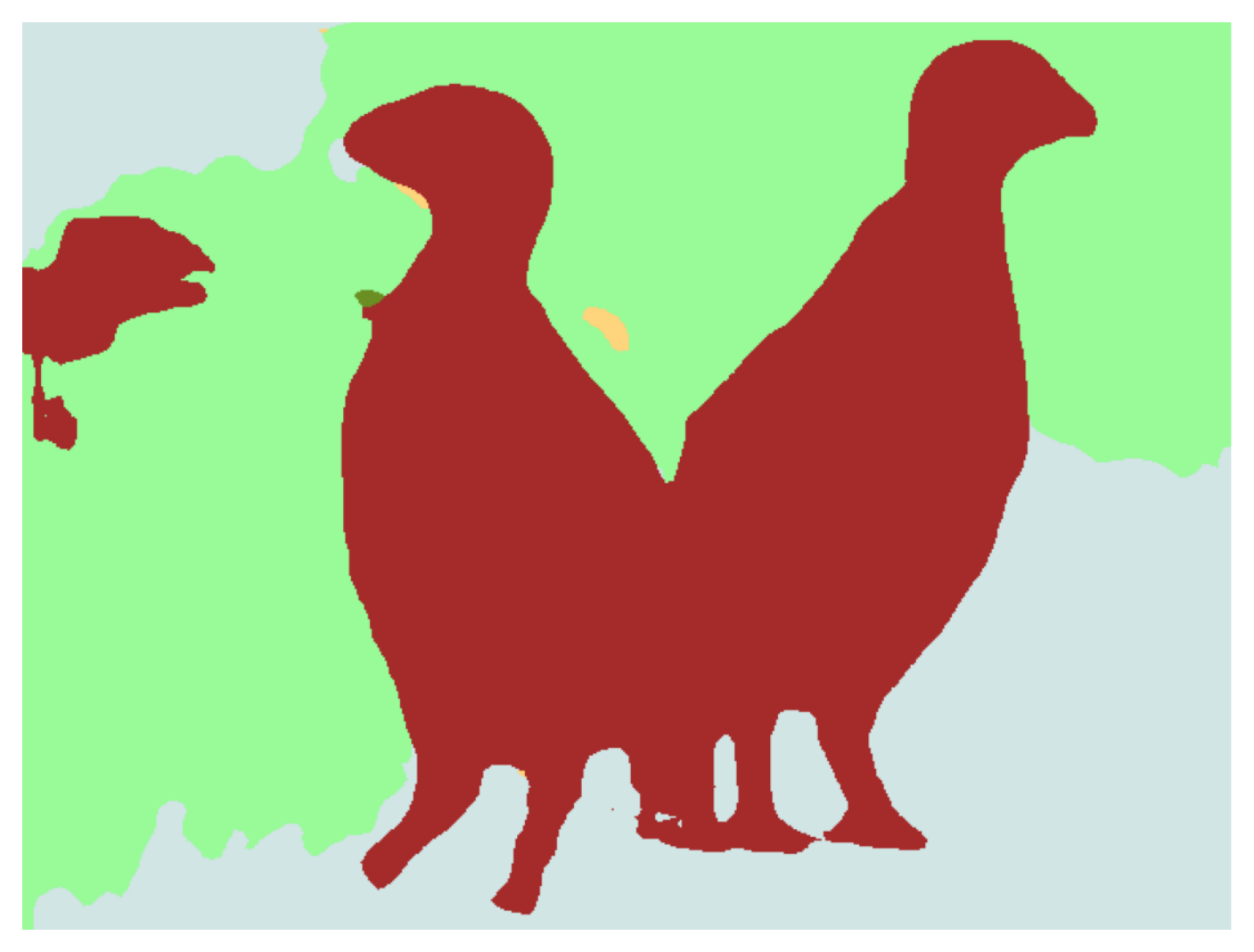}&%
\includegraphics[width=0.3\linewidth]{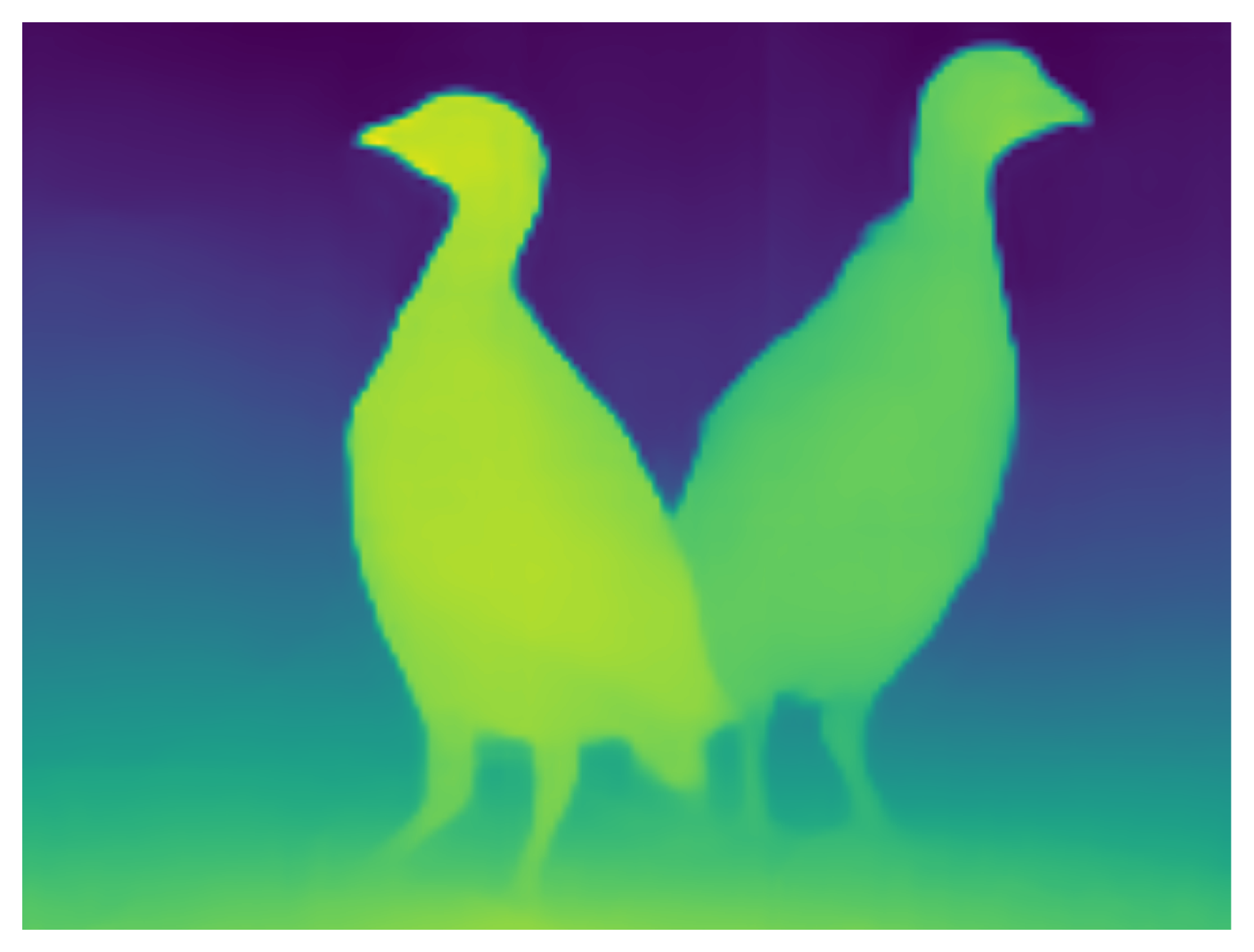}\\%
\includegraphics[width=0.3\linewidth]{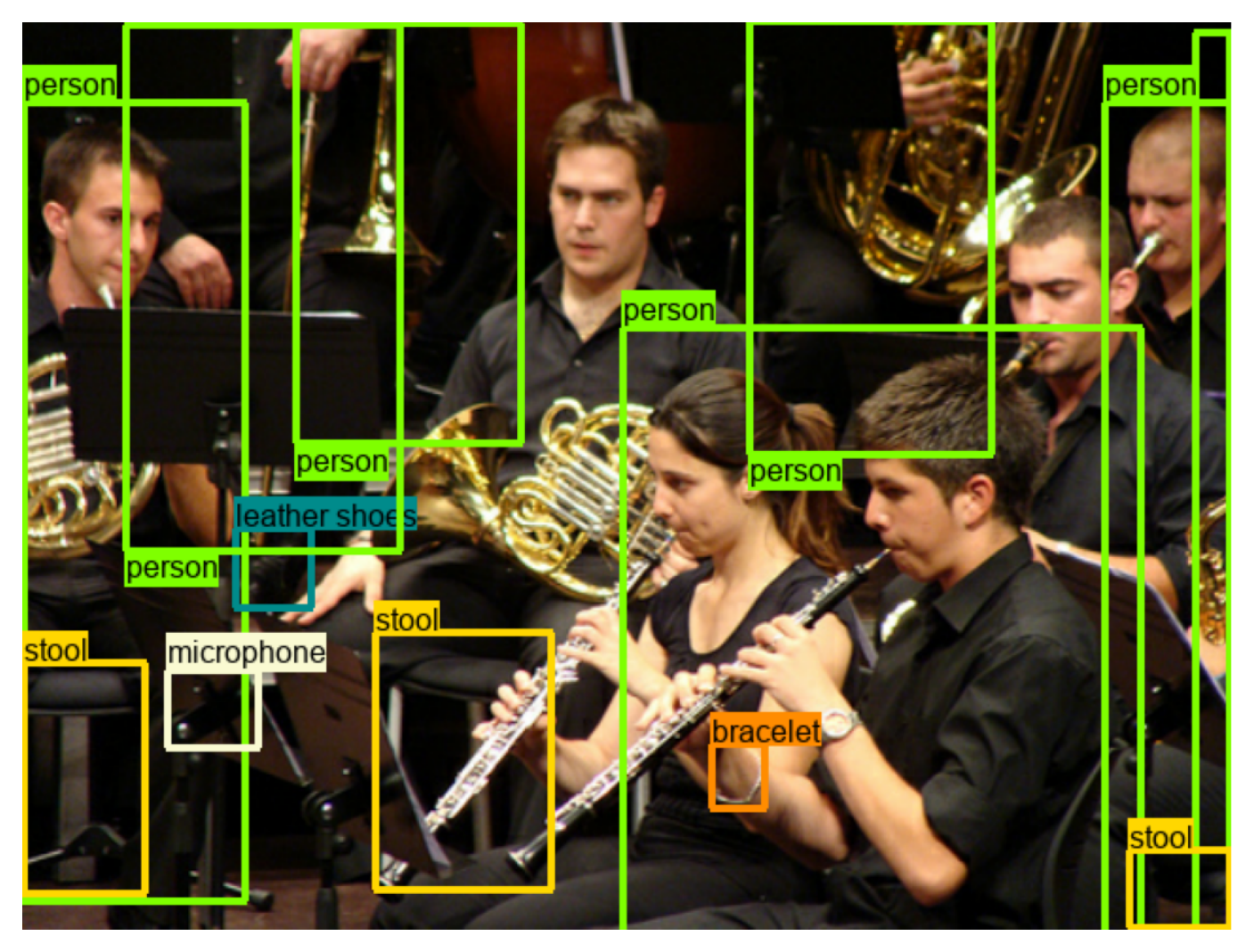}&%
\includegraphics[width=0.3\linewidth]{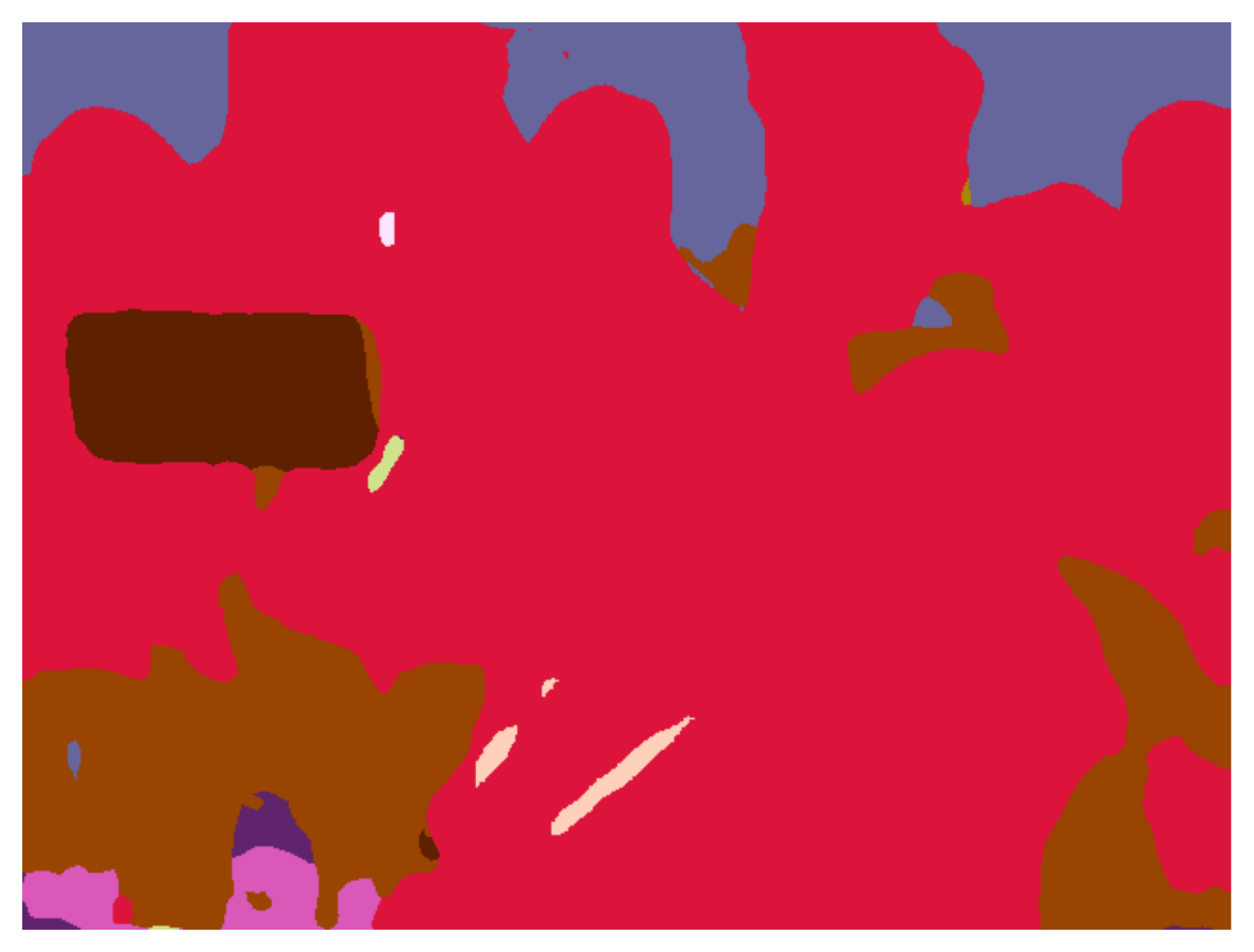}&%
\includegraphics[width=0.3\linewidth]{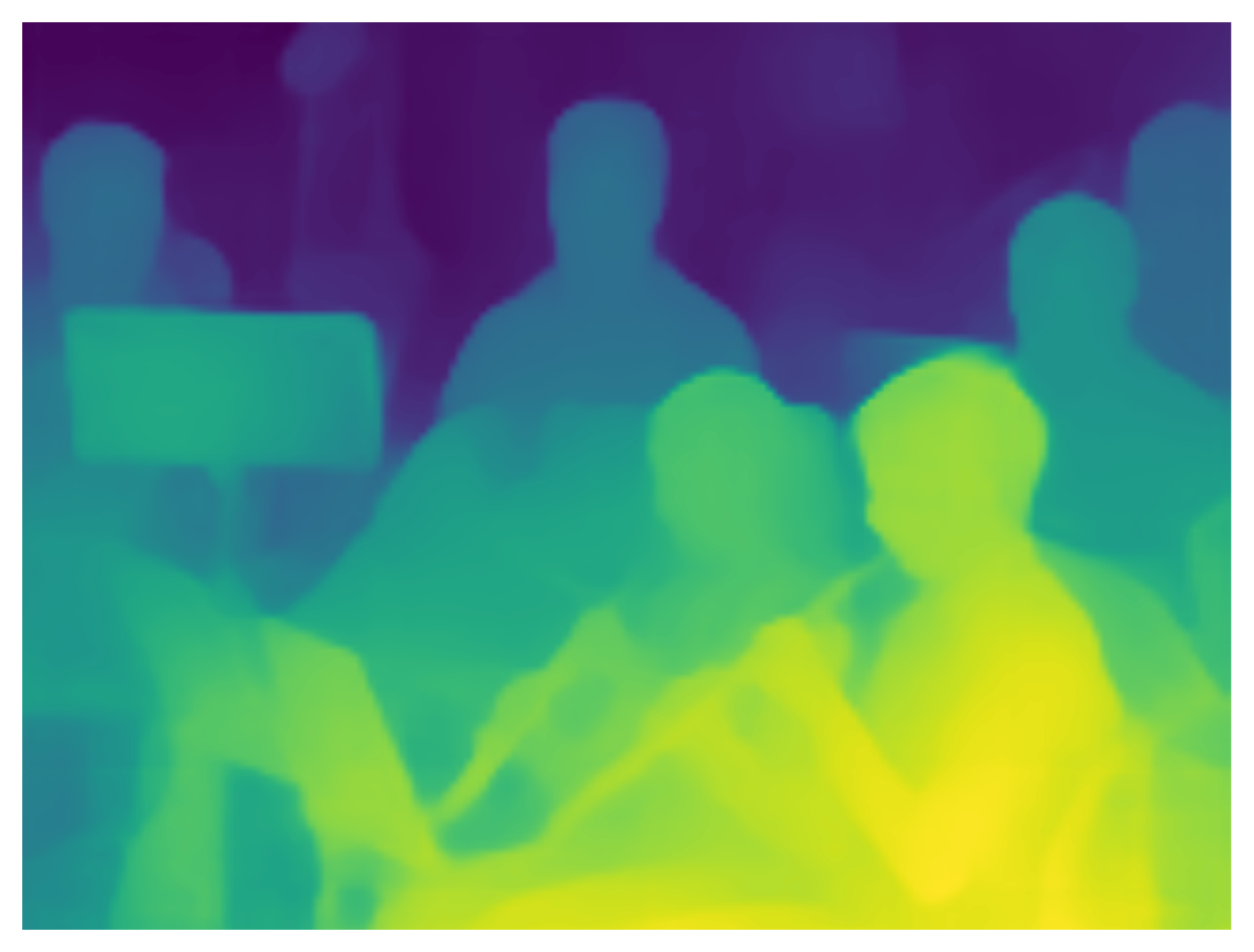}\\%
\includegraphics[width=0.3\linewidth]{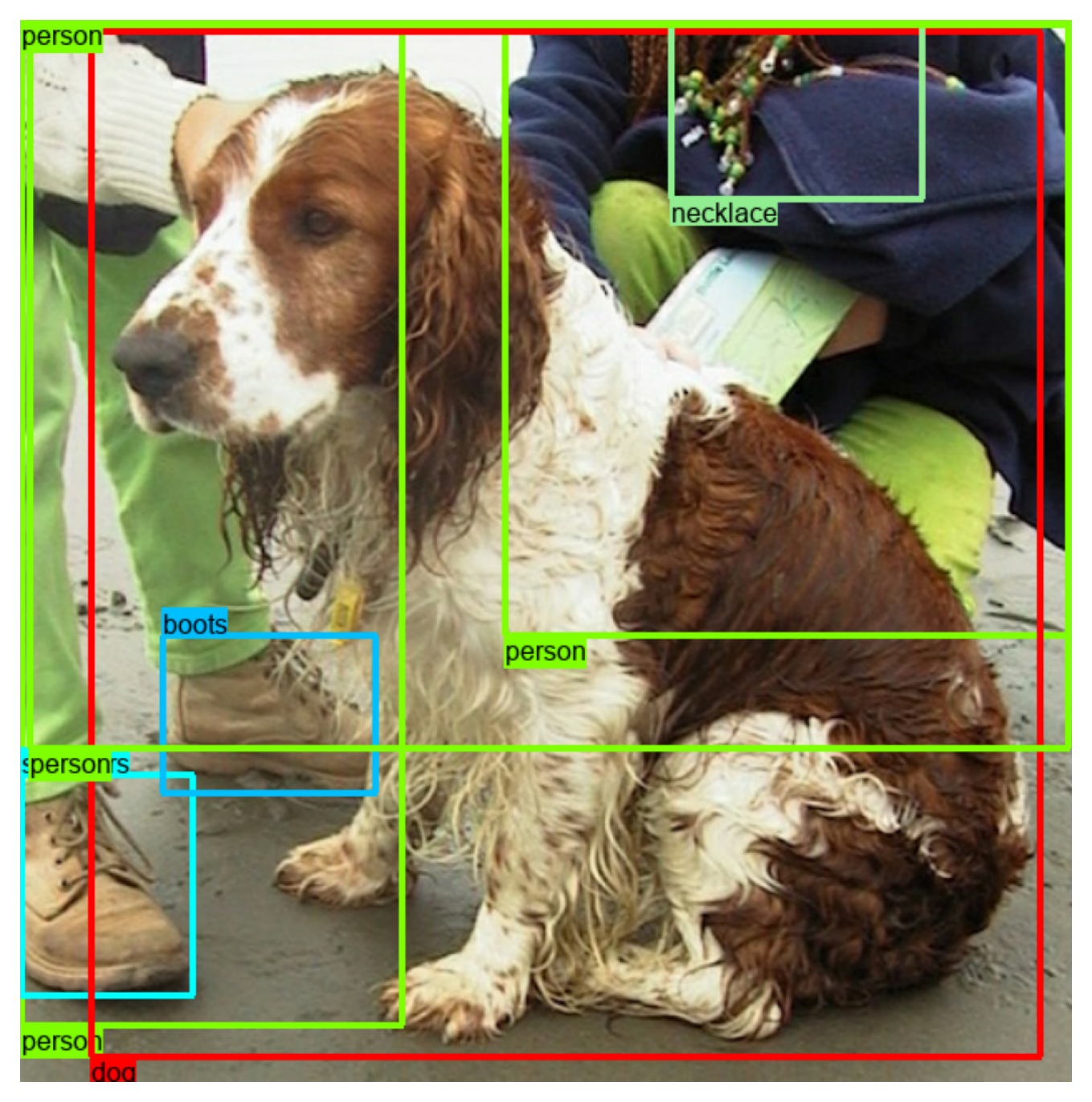}&%
\includegraphics[width=0.3\linewidth]{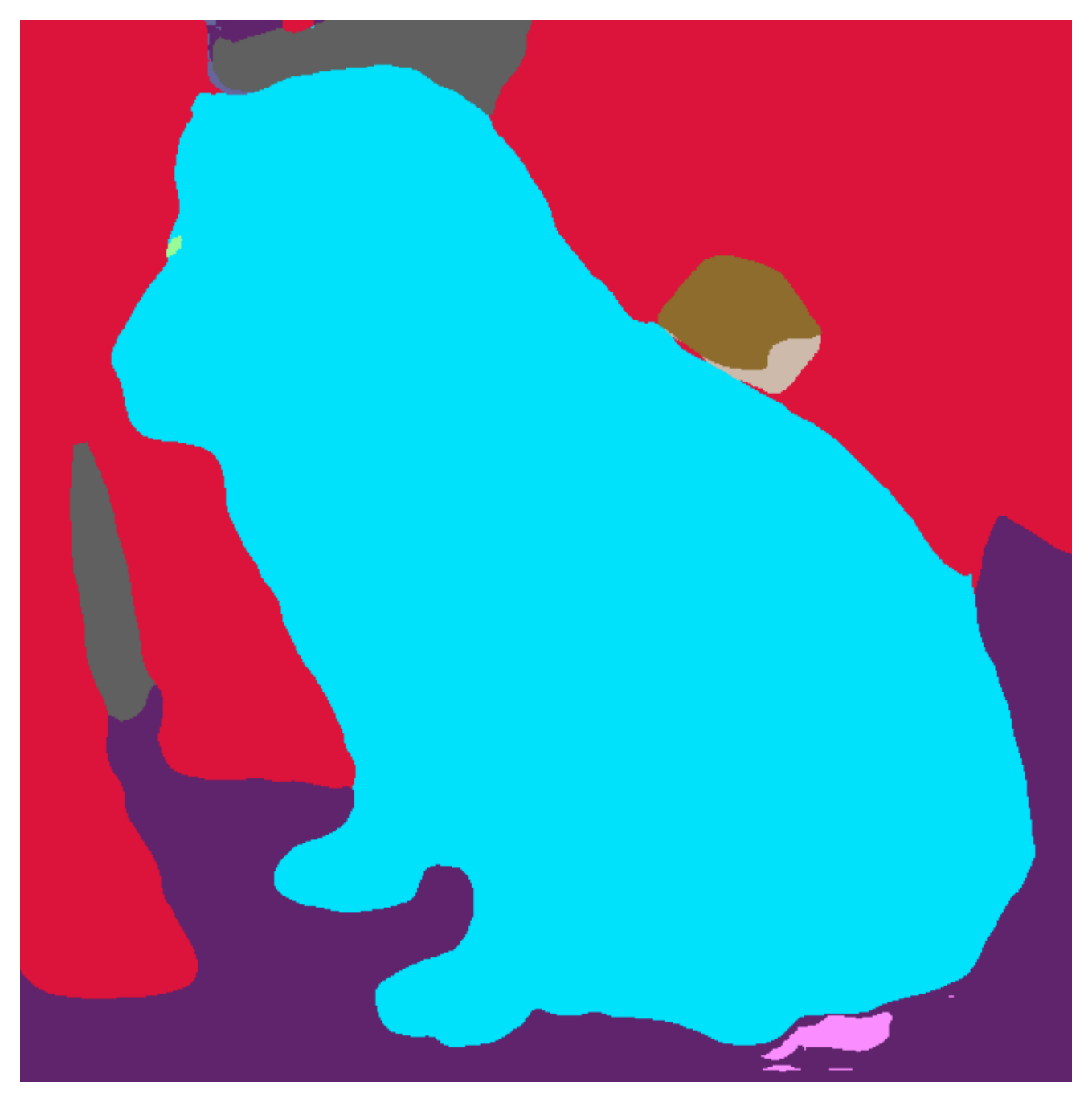}&%
\includegraphics[width=0.3\linewidth]{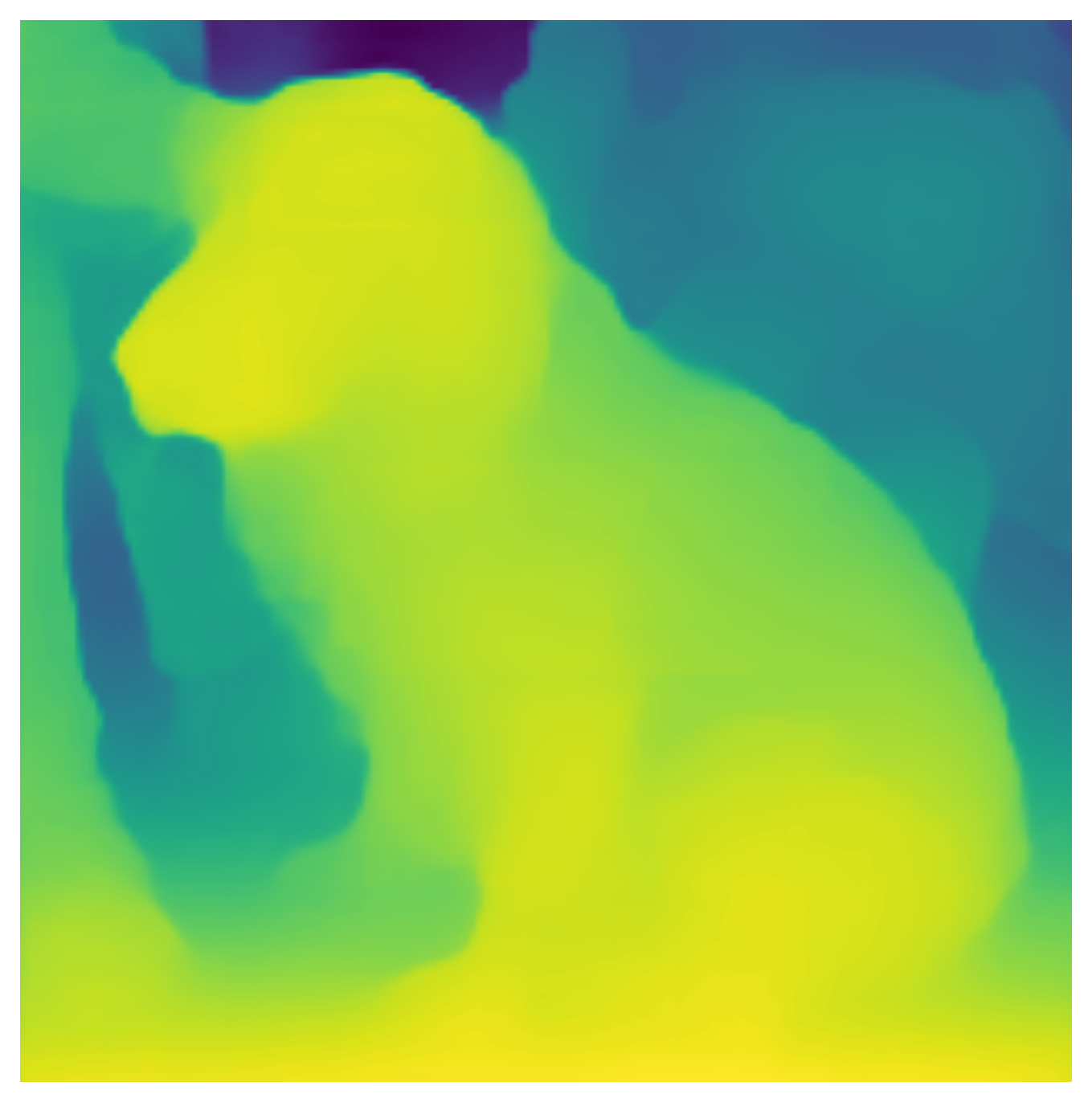}\\%
\includegraphics[width=0.3\linewidth]{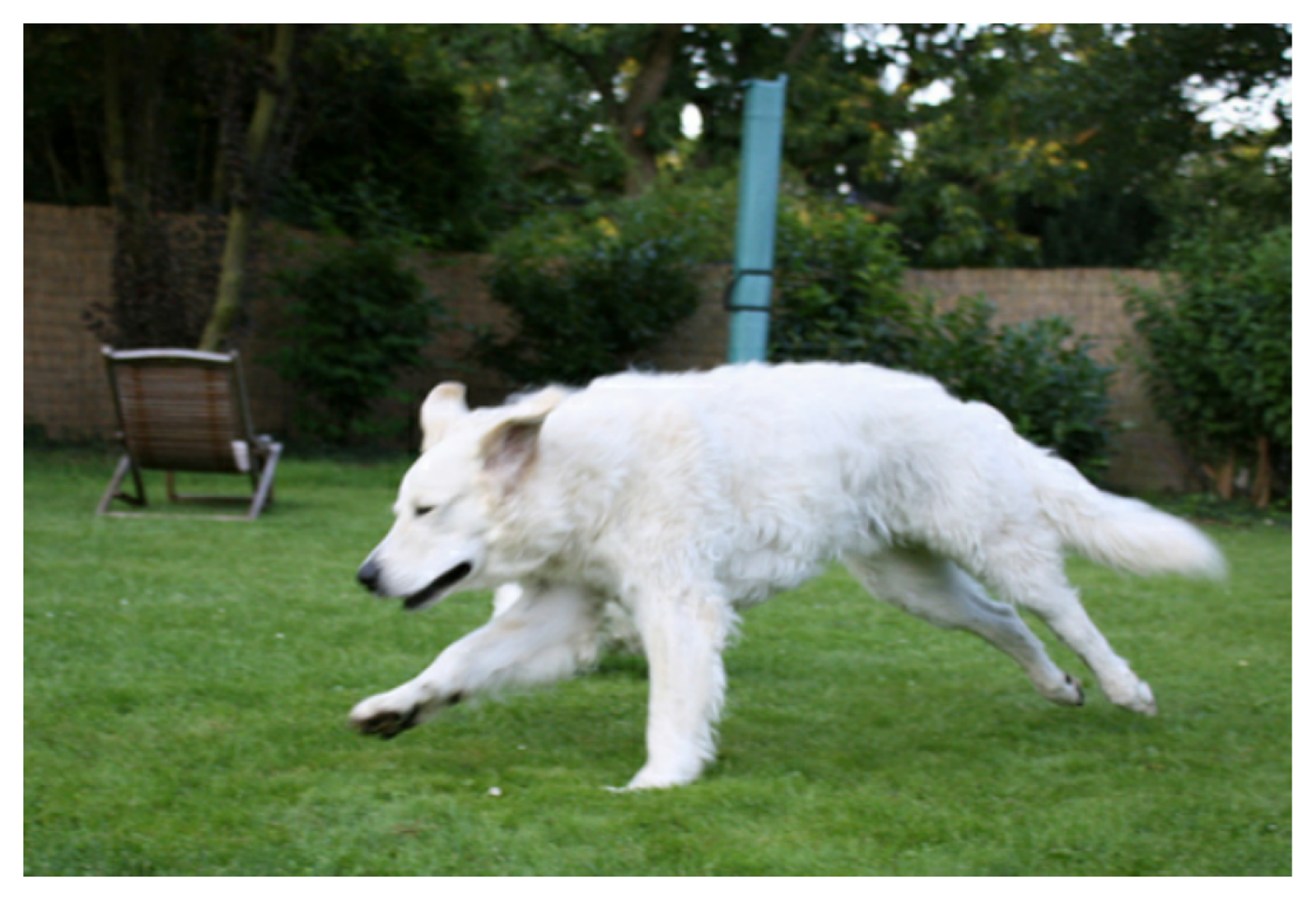}&%
\includegraphics[width=0.3\linewidth]{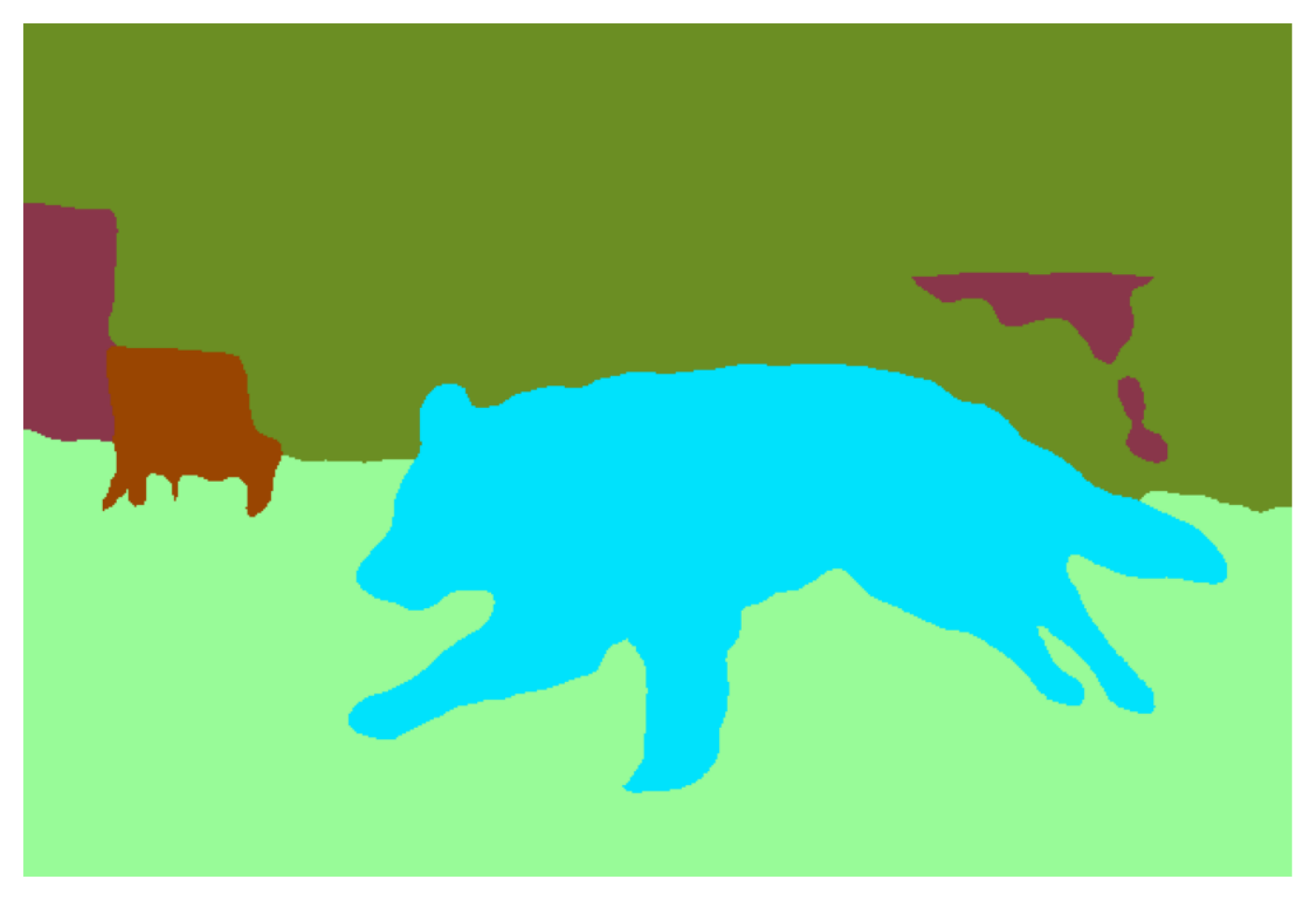}&%
\includegraphics[width=0.3\linewidth]{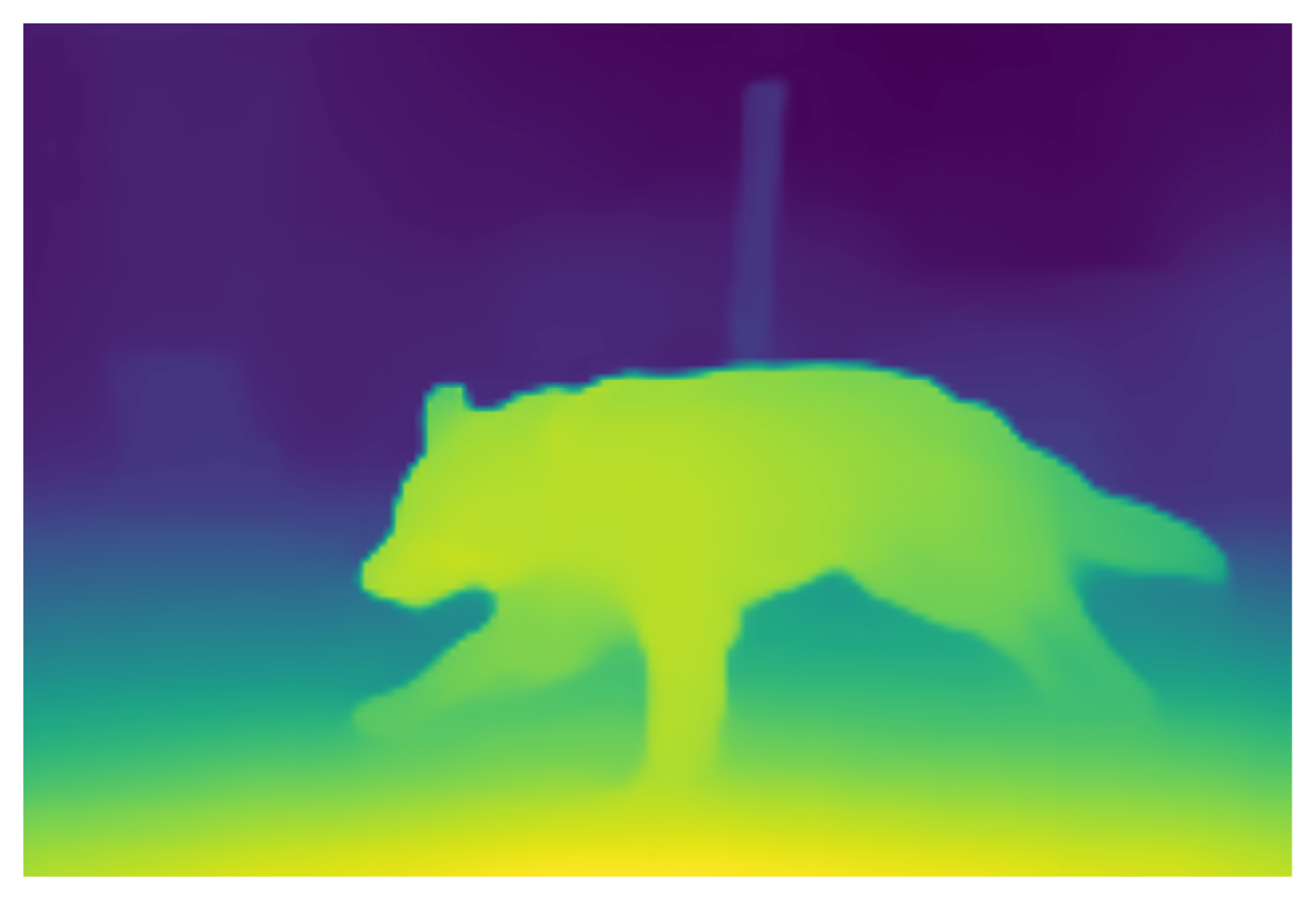}\\%
\includegraphics[width=0.3\linewidth]{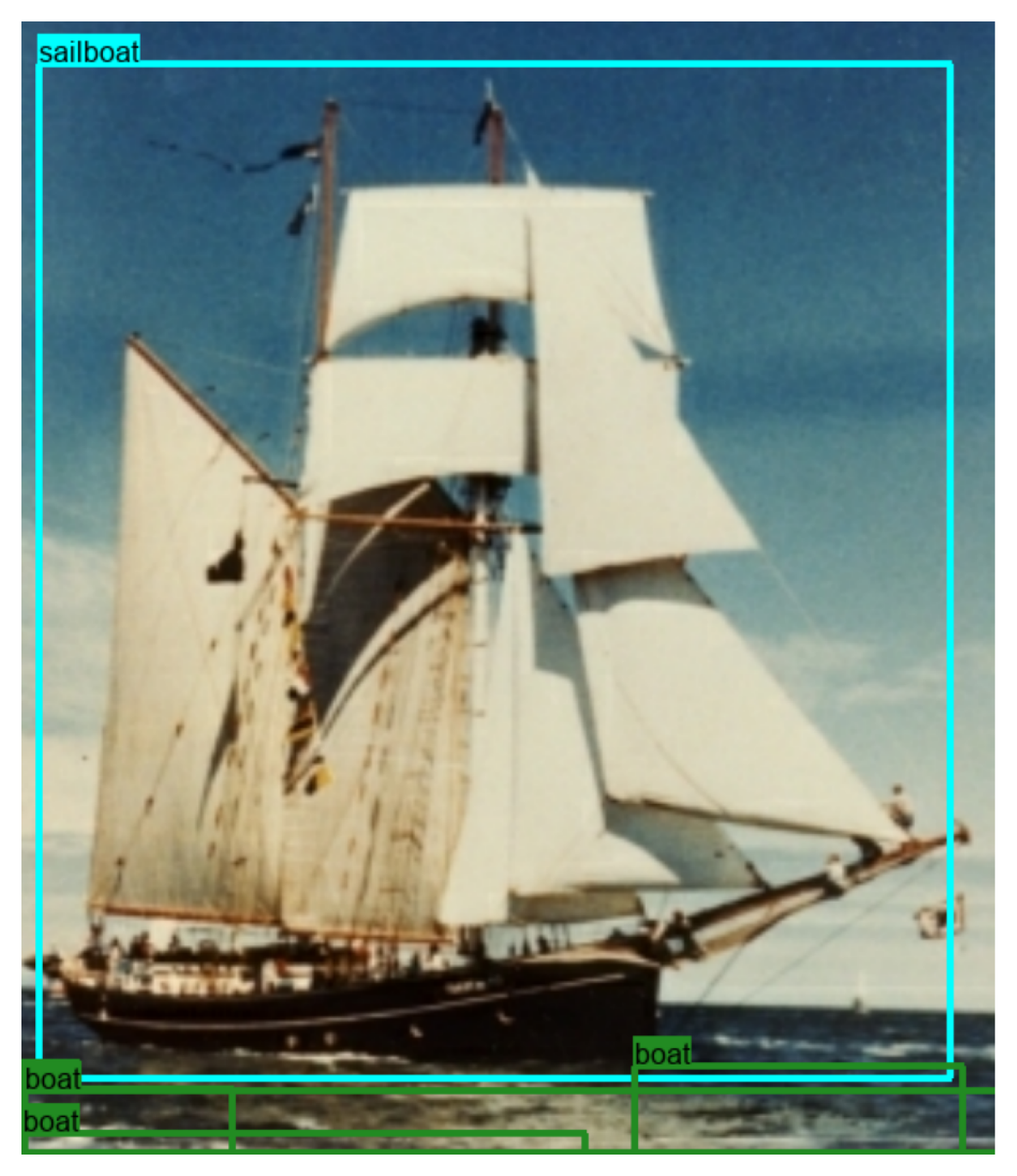}&%
\includegraphics[width=0.3\linewidth]{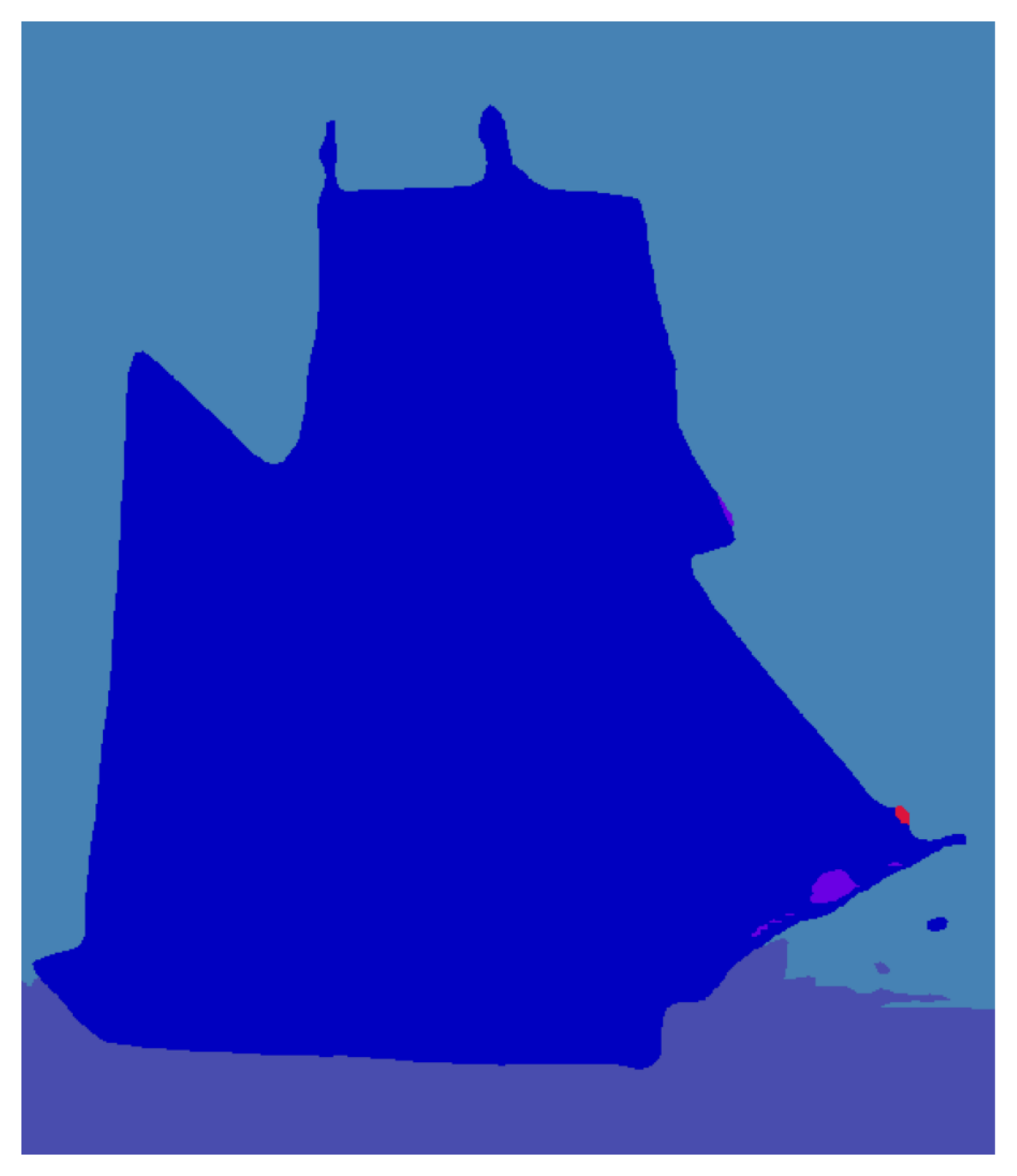}&%
\includegraphics[width=0.3\linewidth]{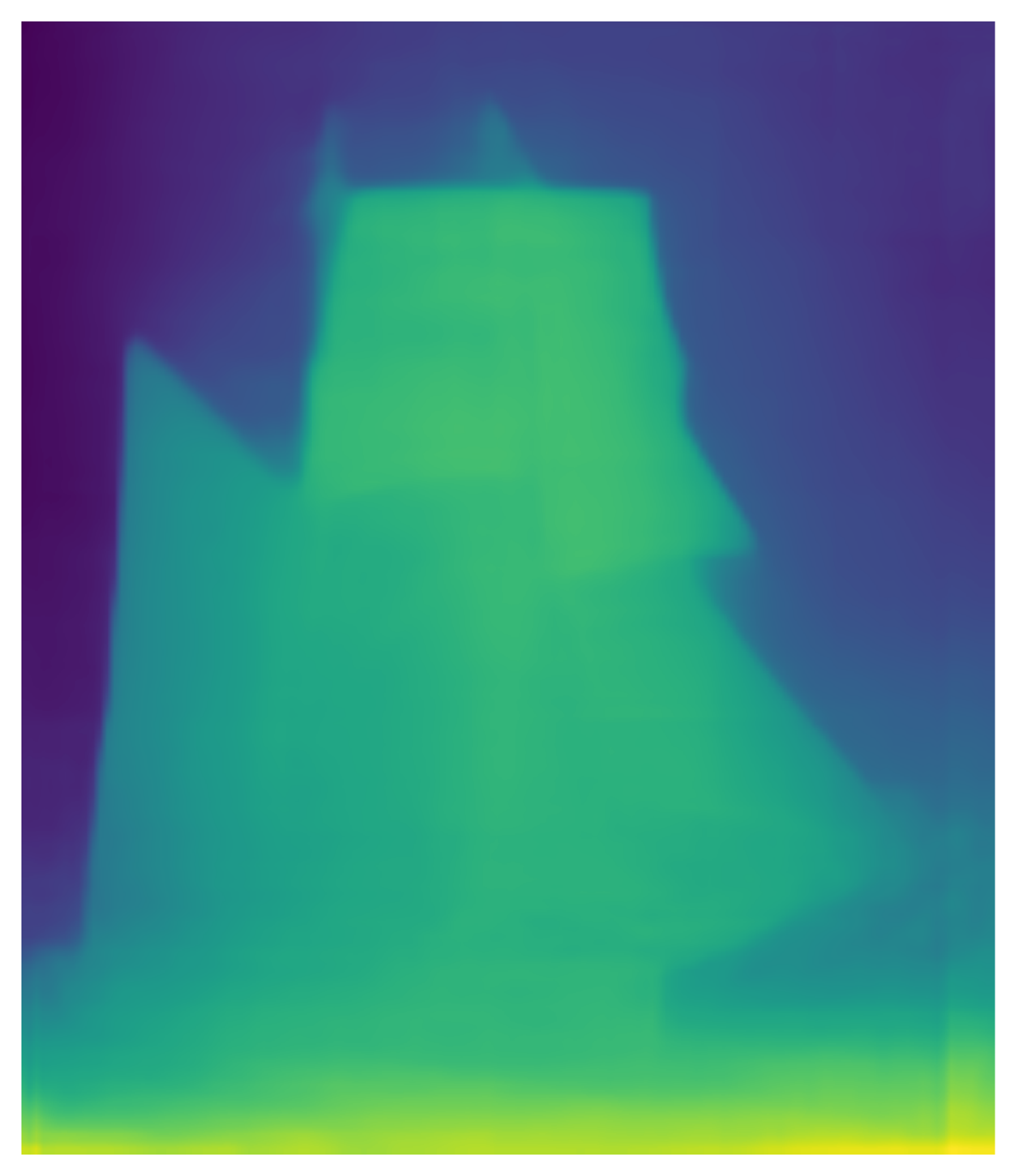}\\%
\end{tabular}%
  \caption{The visualization of inference on ImageNet dataset made by single MuST student model.}%
  \label{fig:imagenet_seg_plabels}%
\end{figure*}

Figure~\ref{fig:imagenet_seg_plabels} shows more visual examples of the predictions made by a single multi-task student model. The images are sampled from the validation set in ImageNet dataset.

\end{appendix}

{\small
\bibliographystyle{ieee_fullname}
\bibliography{egbib}

\begin{thebibliography}{10}\itemsep=-1pt

\bibitem{becker1992self}
Suzanna Becker and Geoffrey~E Hinton.
\newblock Self-organizing neural network that discovers surfaces in random-dot
  stereograms.
\newblock {\em Nature}, 355(6356):161--163, 1992.

\bibitem{caruana1997multitask}
Rich Caruana.
\newblock Multitask learning.
\newblock {\em Machine learning}, 28(1):41--75, 1997.

\bibitem{chen2020naive}
Liang-Chieh Chen, Raphael~Gontijo Lopes, Bowen Cheng, Maxwell~D Collins, Ekin~D
  Cubuk, Barret Zoph, Hartwig Adam, and Jonathon Shlens.
\newblock Naive-student: Leveraging semi-supervised learning in video sequences
  for urban scene segmentation.
\newblock In {\em ECCV}, 2020.

\bibitem{chen2017deeplab}
Liang-Chieh Chen, George Papandreou, Iasonas Kokkinos, Kevin Murphy, and Alan~L
  Yuille.
\newblock Deeplab: Semantic image segmentation with deep convolutional nets,
  atrous convolution, and fully connected crfs.
\newblock {\em IEEE transactions on pattern analysis and machine intelligence},
  40(4):834--848, 2017.

\bibitem{chen2020simple}
Ting Chen, Simon Kornblith, Mohammad Norouzi, and Geoffrey Hinton.
\newblock A simple framework for contrastive learning of visual
  representations.
\newblock In {\em ICML}, 2020.

\bibitem{chen2018gradnorm}
Zhao Chen, Vijay Badrinarayanan, Chen-Yu Lee, and Andrew Rabinovich.
\newblock {G}rad{N}orm: Gradient normalization for adaptive loss balancing in
  deep multitask networks.
\newblock {\em JMLR}, 2018.

\bibitem{chen2020just}
Zhao Chen, Jiquan Ngiam, Yanping Huang, Thang Luong, Henrik Kretzschmar, Yuning
  Chai, and Dragomir Anguelov.
\newblock Just pick a sign: Optimizing deep multitask models with gradient sign
  dropout.
\newblock In {\em NeurIPS}, 2020.

\bibitem{cheng2021maskformer}
Bowen Cheng, Alexander~G. Schwing, and Alexander Kirillov.
\newblock Per-pixel classification is not all you need for semantic
  segmentation.
\newblock {\em CoRR}, abs/2107.06278, 2021.

\bibitem{clark2019bam}
Kevin Clark, Minh{-}Thang Luong, Urvashi Khandelwal, Christopher~D. Manning,
  and Quoc~V. Le.
\newblock Bam! born-again multi-task networks for natural language
  understanding.
\newblock In {\em ACL}, 2019.

\bibitem{cubuk2019randaugment}
Ekin~D. Cubuk, Barret Zoph, Jonathon Shlens, and Quoc~V. Le.
\newblock Randaugment: Practical automated data augmentation with a reduced
  search space, 2019.

\bibitem{dosovitskiy2020image}
Alexey Dosovitskiy, Lucas Beyer, Alexander Kolesnikov, Dirk Weissenborn,
  Xiaohua Zhai, Thomas Unterthiner, Mostafa Dehghani, Matthias Minderer, Georg
  Heigold, Sylvain Gelly, Jakob Uszkoreit, and Neil Houlsby.
\newblock An image is worth 16x16 words: Transformers for image recognition at
  scale, 2020.

\bibitem{du2020selftraining}
Jingfei Du, Edouard Grave, Beliz Gunel, Vishrav Chaudhary, Onur Celebi, Michael
  Auli, Ves Stoyanov, and Alexis Conneau.
\newblock Self-training improves pre-training for natural language
  understanding, 2020.

\bibitem{everingham2010pascal}
Mark Everingham, Luc Van~Gool, Christopher~KI Williams, John Winn, and Andrew
  Zisserman.
\newblock The pascal visual object classes (voc) challenge.
\newblock {\em IJCV}, 2010.

\bibitem{ghiasi2021copypaste}
Golnaz Ghiasi, Yin Cui, Aravind Srinivas, Rui Qian, Tsung-Yi Lin, Ekin~D.
  Cubuk, Quoc~V. Le, and Barret Zoph.
\newblock Simple copy-paste is a strong data augmentation method for instance
  segmentation.
\newblock In {\em Proceedings of the IEEE/CVF Conference on Computer Vision and
  Pattern Recognition (CVPR)}, pages 2918--2928, June 2021.

\bibitem{girshick2015fast}
Ross Girshick.
\newblock Fast r-cnn.
\newblock In {\em ICCV}, 2015.

\bibitem{priya2017scaling}
Priya Goyal, Piotr Doll{\'{a}}r, Ross~B. Girshick, Pieter Noordhuis, Lukasz
  Wesolowski, Aapo Kyrola, Andrew Tulloch, Yangqing Jia, and Kaiming He.
\newblock Accurate, large minibatch {SGD:} training imagenet in 1 hour.
\newblock {\em arXiv preprint arXiv:1706.02677}, 2017.

\bibitem{grill2020bootstrap}
Jean-Bastien Grill, Florian Strub, Florent Altché, Corentin Tallec, Pierre~H.
  Richemond, Elena Buchatskaya, Carl Doersch, Bernardo~Avila Pires,
  Zhaohan~Daniel Guo, Mohammad~Gheshlaghi Azar, Bilal Piot, Koray Kavukcuoglu,
  Rémi Munos, and Michal Valko.
\newblock Bootstrap your own latent: A new approach to self-supervised
  learning.
\newblock In {\em NeurIPS}, 2020.

\bibitem{he2019momentum}
Kaiming He, Haoqi Fan, Yuxin Wu, Saining Xie, and Ross Girshick.
\newblock Momentum contrast for unsupervised visual representation learning.
\newblock In {\em CVPR}, 2020.

\bibitem{rethinking}
Kaiming He, Ross Girshick, and Piotr Doll{\'a}r.
\newblock Rethinking imagenet pre-training.
\newblock In {\em ICCV}, 2019.

\bibitem{mrcnn}
Kaiming He, Georgia Gkioxari, Piotr Doll{\'a}r, and Ross Girshick.
\newblock Mask r-cnn.
\newblock In {\em ICCV}, 2017.

\bibitem{resnet}
Kaiming He, Xiangyu Zhang, Shaoqing Ren, and Jian Sun.
\newblock Deep residual learning for image recognition.
\newblock In {\em CVPR}, 2016.

\bibitem{henaff2019data}
Olivier~J H{\'e}naff, Aravind Srinivas, Jeffrey De~Fauw, Ali Razavi, Carl
  Doersch, SM Eslami, and Aaron van~den Oord.
\newblock Data-efficient image recognition with contrastive predictive coding.
\newblock {\em arXiv preprint arXiv:1905.09272}, 2019.

\bibitem{hinton2015distilling}
Geoffrey Hinton, Oriol Vinyals, and Jeff Dean.
\newblock Distilling the knowledge in a neural network.
\newblock {\em arXiv preprint arXiv:1503.02531}, 2015.

\bibitem{chao2021align}
Chao Jia, Yinfei Yang, Ye Xia, Yi{-}Ting Chen, Zarana Parekh, Hieu Pham,
  Quoc~V. Le, Yun{-}Hsuan Sung, Zhen Li, and Tom Duerig.
\newblock Scaling up visual and vision-language representation learning with
  noisy text supervision.
\newblock {\em CoRR}, abs/2102.05918, 2021.

\bibitem{jing2020self}
Longlong Jing and Yingli Tian.
\newblock Self-supervised visual feature learning with deep neural networks: A
  survey.
\newblock {\em IEEE Transactions on Pattern Analysis and Machine Intelligence},
  2020.

\bibitem{kendall2017multi}
Alex Kendall, Yarin Gal, and Roberto Cipolla.
\newblock Multi-task learning using uncertainty to weigh losses for scene
  geometry and semantics.
\newblock In {\em Proceedings of the IEEE Conference on Computer Vision and
  Pattern Recognition ({CVPR})}, 2018.

\bibitem{kim2018deep}
Youngjung Kim, Hyungjoo Jung, Dongbo Min, and Kwanghoon Sohn.
\newblock Deep monocular depth estimation via integration of global and local
  predictions.
\newblock {\em IEEE transactions on Image Processing}, 27(8):4131--4144, 2018.

\bibitem{Kirillov_2019_CVPR}
Alexander Kirillov, Kaiming He, Ross Girshick, Carsten Rother, and Piotr
  Dollar.
\newblock Panoptic segmentation.
\newblock In {\em CVPR}, June 2019.

\bibitem{Kokkinos_2017_CVPR}
Iasonas Kokkinos.
\newblock Ubernet: Training a universal convolutional neural network for low-,
  mid-, and high-level vision using diverse datasets and limited memory.
\newblock In {\em CVPR}, 2017.

\bibitem{kolesnikov2020big}
Alexander Kolesnikov, Lucas Beyer, Xiaohua Zhai, Joan Puigcerver, Jessica Yung,
  Sylvain Gelly, and Neil Houlsby.
\newblock Big transfer (bit): General visual representation learning.
\newblock In {\em ECCV}, 2020.

\bibitem{cifar}
Alex Krizhevsky.
\newblock Learning multiple layers of features from tiny images.
\newblock Technical report, 2009.

\bibitem{OpenImages}
Alina Kuznetsova, Hassan Rom, Neil Alldrin, Jasper Uijlings, Ivan Krasin, Jordi
  Pont-Tuset, Shahab Kamali, Stefan Popov, Matteo Malloci, Alexander
  Kolesnikov, Tom Duerig, and Vittorio Ferrari.
\newblock The open images dataset v4: Unified image classification, object
  detection, and visual relationship detection at scale.
\newblock {\em IJCV}, 2020.

\bibitem{lee2013pseudo}
Dong-Hyun Lee et~al.
\newblock Pseudo-label: The simple and efficient semi-supervised learning
  method for deep neural networks.
\newblock In {\em Workshop on challenges in representation learning, ICML},
  2013.

\bibitem{li2019detection_pretraining}
Hengduo Li, Bharat Singh, Mahyar Najibi, Zuxuan Wu, and Larry~S. Davis.
\newblock An analysis of pre-training on object detection.
\newblock {\em CoRR}, abs/1904.05871, 2019.

\bibitem{li2018megadepth}
Zhengqi Li and Noah Snavely.
\newblock Megadepth: Learning single-view depth prediction from internet
  photos.
\newblock In {\em CVPR}, 2018.

\bibitem{fpn}
Tsung-Yi Lin, Piotr Doll{\'a}r, Ross Girshick, Kaiming He, Bharath Hariharan,
  and Serge Belongie.
\newblock Feature pyramid networks for object detection.
\newblock In {\em CVPR}, 2017.

\bibitem{coco}
Tsung-Yi Lin, Michael Maire, Serge Belongie, James Hays, Pietro Perona, Deva
  Ramanan, Piotr Doll{\'a}r, and C~Lawrence Zitnick.
\newblock Microsoft coco: Common objects in context.
\newblock In {\em ECCV}, 2014.

\bibitem{gao2019multitasknlp}
Xiaodong Liu, Pengcheng He, Weizhu Chen, and Jianfeng Gao.
\newblock Improving multi-task deep neural networks via knowledge distillation
  for natural language understanding.
\newblock In {\em ACL}, 2019.

\bibitem{Mahajan_2018_ECCV}
Dhruv Mahajan, Ross Girshick, Vignesh Ramanathan, Kaiming He, Manohar Paluri,
  Yixuan Li, Ashwin Bharambe, and Laurens van~der Maaten.
\newblock Exploring the limits of weakly supervised pretraining.
\newblock In {\em ECCV}, September 2018.

\bibitem{misra2020self}
Ishan Misra and Laurens van~der Maaten.
\newblock Self-supervised learning of pretext-invariant representations.
\newblock In {\em CVPR}, 2020.

\bibitem{purushwalkam2020demystifying}
Senthil Purushwalkam and Abhinav Gupta.
\newblock Demystifying contrastive self-supervised learning: Invariances,
  augmentations and dataset biases.
\newblock {\em arXiv preprint arXiv:2007.13916}, 2020.

\bibitem{raffel2019exploring}
Colin Raffel, Noam Shazeer, Adam Roberts, Katherine Lee, Sharan Narang, Michael
  Matena, Yanqi Zhou, Wei Li, and Peter~J Liu.
\newblock Exploring the limits of transfer learning with a unified text-to-text
  transformer.
\newblock {\em JMLR}, 2020.

\bibitem{ranftl2021dpt}
Ren\'{e} Ranftl, Alexey Bochkovskiy, and Vladlen Koltun.
\newblock Vision transformers for dense prediction.
\newblock {\em ArXiv preprint}, 2021.

\bibitem{Ranftl2020}
Ren\'{e} Ranftl, Katrin Lasinger, David Hafner, Konrad Schindler, and Vladlen
  Koltun.
\newblock Towards robust monocular depth estimation: Mixing datasets for
  zero-shot cross-dataset transfer.
\newblock {\em IEEE Transactions on Pattern Analysis and Machine Intelligence
  (TPAMI)}, 2020.

\bibitem{rosenberg2005semi}
Chuck Rosenberg, Martial Hebert, and Henry Schneiderman.
\newblock Semi-supervised self-training of object detection models.
\newblock In {\em IEEE Workshops on Applications of Computer Vision
  (WACV/MOTION'05)}, 2005.

\bibitem{ruder2017overview}
Sebastian Ruder.
\newblock An overview of multi-task learning in deep neural networks.
\newblock {\em arXiv preprint arXiv:1706.05098}, 2017.

\bibitem{ilsvrc}
Olga Russakovsky, Jia Deng, Hao Su, Jonathan Krause, Sanjeev Satheesh, Sean Ma,
  Zhiheng Huang, Andrej Karpathy, Aditya Khosla, Michael Bernstein, et~al.
\newblock Imagenet large scale visual recognition challenge.
\newblock {\em IJCV}, 2015.

\bibitem{scudder1965probability}
H Scudder.
\newblock Probability of error of some adaptive pattern-recognition machines.
\newblock {\em IEEE Transactions on Information Theory}, 11(3):363--371, 1965.

\bibitem{objects365}
Shuai Shao, Zeming Li, Tianyuan Zhang, Chao Peng, Gang Yu, Xiangyu Zhang, Jing
  Li, and Jian Sun.
\newblock Objects365: A large-scale, high-quality dataset for object detection.
\newblock In {\em ICCV}, 2019.

\bibitem{silberman2012indoor}
Nathan Silberman, Derek Hoiem, Pushmeet Kohli, and Rob Fergus.
\newblock Indoor segmentation and support inference from rgbd images.
\newblock In {\em ECCV}, 2012.

\bibitem{sun2017revisiting}
Chen Sun, Abhinav Shrivastava, Saurabh Singh, and Abhinav Gupta.
\newblock Revisiting unreasonable effectiveness of data in deep learning era.
\newblock In {\em ICCV}, 2017.

\bibitem{tan19enet}
Mingxing Tan and Quoc Le.
\newblock {E}fficient{N}et: Rethinking model scaling for convolutional neural
  networks.
\newblock In Kamalika Chaudhuri and Ruslan Salakhutdinov, editors, {\em
  Proceedings of the 36th International Conference on Machine Learning},
  volume~97 of {\em Proceedings of Machine Learning Research}, pages
  6105--6114. PMLR, 09--15 Jun 2019.

\bibitem{tian2019contrastive}
Yonglong Tian, Dilip Krishnan, and Phillip Isola.
\newblock Contrastive multiview coding.
\newblock In {\em ECCV}, 2019.

\bibitem{vasiljevic2019diode}
Igor Vasiljevic, Nick Kolkin, Shanyi Zhang, Ruotian Luo, Haochen Wang, Falcon~Z
  Dai, Andrea~F Daniele, Mohammadreza Mostajabi, Steven Basart, Matthew~R
  Walter, et~al.
\newblock Diode: A dense indoor and outdoor depth dataset.
\newblock {\em arXiv preprint arXiv:1908.00463}, 2019.

\bibitem{wang2019web}
Chaoyang Wang, Simon Lucey, Federico Perazzi, and Oliver Wang.
\newblock Web stereo video supervision for depth prediction from dynamic
  scenes.
\newblock In {\em 2019 International Conference on 3D Vision (3DV)}, pages
  348--357. IEEE, 2019.

\bibitem{xian2018monocular}
Ke Xian, Chunhua Shen, Zhiguo Cao, Hao Lu, Yang Xiao, Ruibo Li, and Zhenbo Luo.
\newblock Monocular relative depth perception with web stereo data supervision.
\newblock In {\em CVPR}, 2018.

\bibitem{xiao2018unified}
Tete Xiao, Yingcheng Liu, Bolei Zhou, Yuning Jiang, and Jian Sun.
\newblock Unified perceptual parsing for scene understanding.
\newblock In {\em European Conference on Computer Vision}. Springer, 2018.

\bibitem{xie2019self}
Qizhe Xie, Eduard Hovy, Minh-Thang Luong, and Quoc~V Le.
\newblock Self-training with noisy student improves imagenet classification.
\newblock In {\em CVPR}, 2020.

\bibitem{xu2020selftraining}
Qiantong Xu, Alexei Baevski, Tatiana Likhomanenko, Paden Tomasello, Alexis
  Conneau, Ronan Collobert, Gabriel Synnaeve, and Michael Auli.
\newblock Self-training and pre-training are complementary for speech
  recognition, 2020.

\bibitem{xue2020mt5}
Linting Xue, Noah Constant, Adam Roberts, Mihir Kale, Rami Al-Rfou, Aditya
  Siddhant, Aditya Barua, and Colin Raffel.
\newblock mt5: A massively multilingual pre-trained text-to-text transformer.
\newblock In {\em NAACL}, 2020.

\bibitem{zeki2019billion}
I.~Zeki Yalniz, Herv{\'{e}} J{\'{e}}gou, Kan Chen, Manohar Paluri, and Dhruv
  Mahajan.
\newblock Billion-scale semi-supervised learning for image classification.
\newblock {\em CoRR}, abs/1905.00546, 2019.

\bibitem{yarowsky1995unsupervised}
David Yarowsky.
\newblock Unsupervised word sense disambiguation rivaling supervised methods.
\newblock In {\em ACL}, 1995.

\bibitem{yu2020gradient}
Tianhe Yu, Saurabh Kumar, Abhishek Gupta, Sergey Levine, Karol Hausman, and
  Chelsea Finn.
\newblock Gradient surgery for multi-task learning.
\newblock {\em arXiv preprint arXiv:2001.06782}, 2020.

\bibitem{zamir2018taskonomy}
Amir~R. Zamir, Alexander Sax, William Shen, Leonidas~J. Guibas, Jitendra Malik,
  and Silvio Savarese.
\newblock Taskonomy: Disentangling task transfer learning.
\newblock In {\em CVPR}, 2018.

\bibitem{zhang2020pushing}
Yu Zhang, James Qin, Daniel~S. Park, Wei Han, Chung-Cheng Chiu, Ruoming Pang,
  Quoc~V. Le, and Yonghui Wu.
\newblock Pushing the limits of semi-supervised learning for automatic speech
  recognition, 2020.

\bibitem{zhou2019semantic}
Bolei Zhou, Hang Zhao, Xavier Puig, Tete Xiao, Sanja Fidler, Adela Barriuso,
  and Antonio Torralba.
\newblock Semantic understanding of scenes through the ade20k dataset.
\newblock {\em IJCV}, 127(3):302--321, 2019.

\bibitem{zoph2020rethink}
Barret Zoph, Golnaz Ghiasi, Tsung-Yi Lin, Yin Cui, Hanxiao Liu, Ekin~D. Cubuk,
  and Quoc~V. Le.
\newblock Rethinking pre-training and self-training.
\newblock In {\em NeurIPS}, 2020.

\end{thebibliography}
}

\end{document}